\documentclass[11pt]{article}

\usepackage[english]{babel}

\usepackage[letterpaper,top=2cm,bottom=2cm,left=3cm,right=3cm,marginparwidth=1.75cm]{geometry}

\usepackage{amsmath}
\usepackage{graphicx}
\usepackage[colorlinks=true, allcolors=blue]{hyperref}
\usepackage{csquotes}
\usepackage[backend=biber,style=ieee]{biblatex}
\usepackage{booktabs}
\usepackage{tabularx}
\usepackage{subcaption}
\usepackage{multirow}

\usepackage{xcolor}
\usepackage{soul}

\usepackage{blindtext} 

\DeclareMathOperator*{\argmin}{argmin}

\DeclareUnicodeCharacter{2212}{-}

\title{Deep hybrid model with satellite imagery: how to combine demand modeling and computer vision for travel behavior analysis?}

\author{
    \small
    Qingyi Wang\textsuperscript{1},
    Shenhao Wang\textsuperscript{2,4,7,{*}},
    Yunhan Zheng\textsuperscript{1,2} \\
    \small
    Hongzhou Lin\textsuperscript{3}, 
    Xiaohu Zhang\textsuperscript{6},
    Jinhua Zhao\textsuperscript{2}, 
    Joan Walker\textsuperscript{5}
    }
    
\date{\footnotesize %
    $^1$Department of Civil and Environmental Engineering, $^2$Department of Urban Studies and Planning, $^3$Computer Science and Artificial Intelligence Laboratory, $^4$Media Lab, Massachusetts Institute of Technology, Cambridge, MA\\
    $^5$Civil and Environmental Engineering, University of California at Berkeley, Berkeley, CA\\
    $^6$Urban Planning and Design, The University of Hong Kong, Hong Kong, China \\ 
    $^7$Department of Urban and Regional Planning, University of Florida, Gainesville, FL\\
}

\begin{document}
\maketitle
\begin{abstract}
\noindent
Classical demand modeling analyzes travel behavior using only low-dimensional numeric data (i.e. sociodemographics and travel attributes) but not high-dimensional urban imagery. However, travel behavior depends on the factors represented by both numeric data and urban imagery, thus necessitating a synergetic framework to combine them. This study creates a theoretical framework of \textbf{deep hybrid models} consisting of a mixing operator and a behavioral predictor, thus integrating the numeric and imagery data for travel behavior analysis. Empirically, this framework is applied to analyze travel mode choice using the Chicago MyDailyTravel Survey as the numeric inputs and the satellite images as the imagery inputs. We found that deep hybrid models significantly outperform both classical demand models and deep learning models in predicting aggregate and disaggregate travel behavior. The deep hybrid models can reveal spatial clusters with meaningful sociodemographic associations in the latent space. The models can also generate new satellite images that do not exist in reality and compute the corresponding economic information, such as substitution patterns and social welfare. Overall, the deep hybrid models demonstrate the complementarity between the low-dimensional numeric and high-dimensional imagery data and between the traditional demand modeling and recent deep learning. They enrich the family of hybrid demand models by using deep architecture as the latent space and enabling researchers to conduct associative analysis for sociodemographics, travel decisions, and generated satellite imagery. Future research could address the limitations in interpretability, robustness, and transferability, and propose new methods to further enrich the deep hybrid models.
\end{abstract}


\textbf{*} To whom correspondence should be addressed. E-mail: shenhaowang@ufl.edu
\newpage

\section{Introduction}
\noindent 
Demand modeling has been a theoretically rich field widely applied to various travel behavioral analyses. Researchers created the multinomial logit model to capture random utility maximization as a decision mechanism \cite{McFadden1974}, the nested logit model to represent the tree structure of the alternatives \cite{McFadden1978_residential_location}, the mixed logit model to capture the behavioral heterogeneity in preference parameters \cite{McFadden2000_mixed_logit}, and the hybrid demand model to reveal the latent behavioral structure \cite{Walker2002,Ben-Akiva2002HybridChallenges,Vij2016}. These demand models have been applied to analyze car ownership, travel mode choice, adoption of electric vehicles, and destination choice, among many other travel behaviors \cite{Train1980_auto_ownership, Helveston2015, Bergantino2020}. However, the existing demand models use only low-dimensional numeric data, including sociodemographic characteristics and trip attributes, while lacking the capacity to process high-dimensional unstructured data, such as urban imagery. 
Urban imagery has been shown to contain valuable information on built environment, socioeconomic factors, and mobility patterns by recent deep learning research \cite{Jean2016, Ayush2020, Yeh2020}. 
It seems a natural effort to extend the classical demand models to incorporate urban imagery, thus reflecting a more realistic behavioral mechanism and enriching the demand modeling tools.

To operationalize this idea, the key question is \textit{how to integrate the numeric and urban imagery data, leveraging the computational power of deep learning for urban imagery while retaining economic information for practical uses.} On the one hand, demand models have demonstrated that travel decisions depend on travel time, travel cost, income, age, and other numeric data, which facilitates rigorous microeconomic analysis \cite{Small1981_welfare, McFadden1974}. The microeconomic analysis is valuable because it leverages the random utility theory to compute critical economic parameters such as social welfare and substitution patterns of alternatives \cite{Small1981_welfare, Zamparini2016}. However, an exclusive focus on numeric data misses the tremendous opportunities in the recent big data revolution, in which unstructured data, such as urban imagery, accounts for more than 80\% of data growth \cite{Azad2020_unstructured_data}. On the other hand, the deep learning models in the field of urban computing have used urban imagery - typically satellite or street-view images - to predict sociodemographic characteristics, achieving high predictive performance \cite{Jean2016, Gebru2017}. However, urban computing research exclusively focuses on using urban imagery for prediction, which dismisses the practical needs of computing elasticity, social welfare, market shares, and other important economic factors. A pure deep learning approach ignores the sociodemographic and travel-related attributes, but it is implausible that travel decisions do not depend on income, age, and travel costs. Since each of the two research paradigms only partially captures behavioral realism, this dichotomy necessitates an effort to integrate them, thus successfully incorporating urban imagery into decision analysis while retaining certain economic information for practical use.

\begin{figure}[ht!]
\centering
\includegraphics[height=0.3\linewidth]{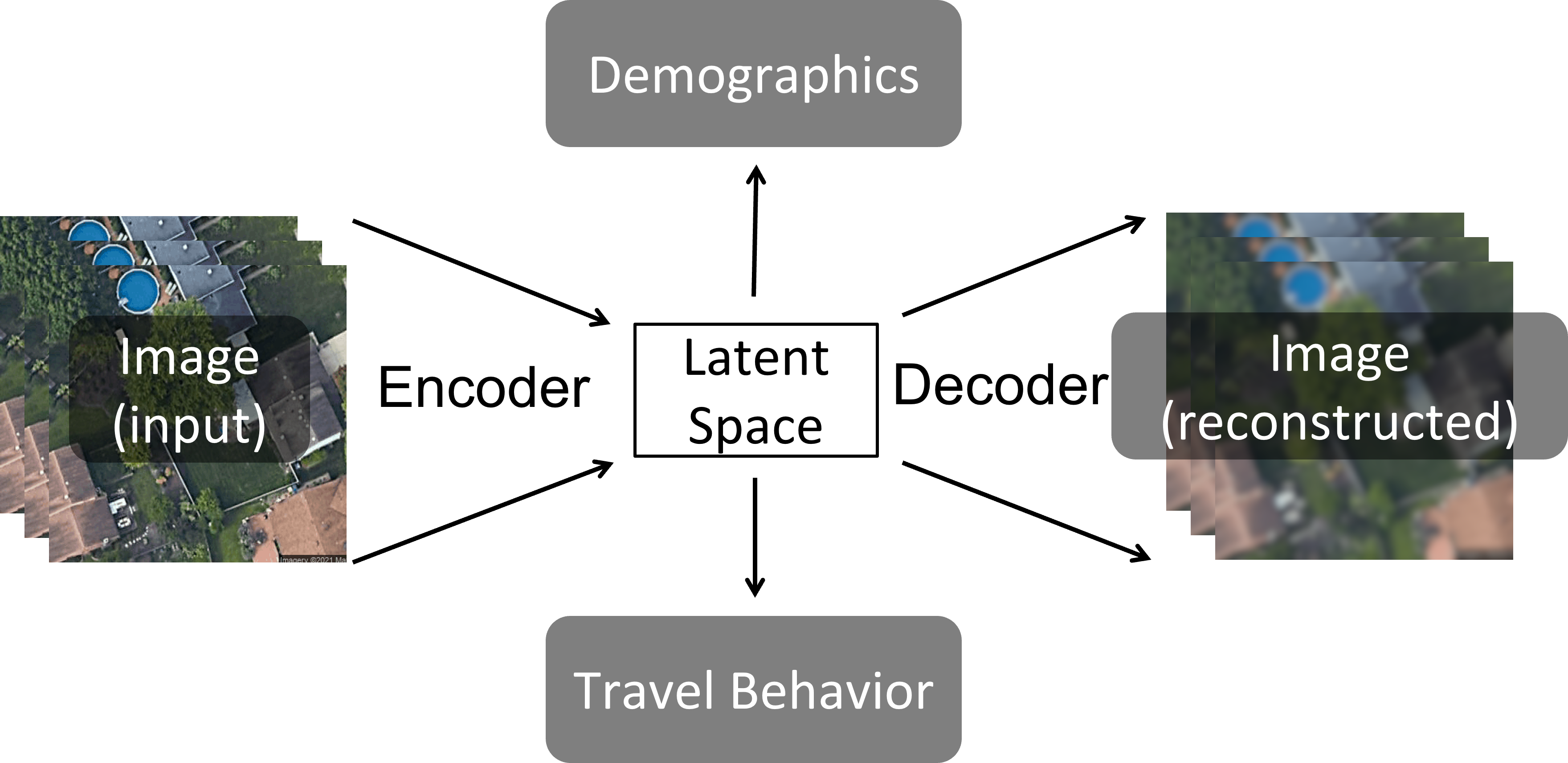}\label{sfig:dhm}
\caption{Diagram of a deep hybrid model (Model 4-6 in Table \ref{tab:mixing_operator_design})}
\label{fig:dhm_framework}
\end{figure}

This study presents a synergetic framework of the deep hybrid model (DHM), which is visually represented by a crossing structure with a vertical and a horizontal axis, as shown in Figure \ref{fig:dhm_framework}. The vertical axis represents the components in classical demand modeling, including the numeric inputs and outputs (e.g., sociodemographics, travel attributes, and behavioral outputs). The horizontal axis represents the components of deep learning, specifically an autoencoder that encodes and regenerates urban imagery. These two axes are connected through a latent space, which serves as the core to integrate the numeric data and urban imagery. This framework is named as ``deep hybrid’’ because it resembles and enriches the classical hybrid demand models \cite{Ben-Akiva2002HybridChallenges}. It resembles the classical hybrid model because it is similar to the visual diagram of Figure 3 in \cite{Ben-Akiva2002HybridChallenges}, with our horizontal axis resembling the measurement model and our vertical axis resembling the structural model. It enriches the classical hybrid demand models with deep architectures to construct a latent space with higher dimensions, which are capable of processing the unstructured data, such as urban imagery. This framework builds upon the recent efforts in deep choice analysis \cite{Wang2020DeepFunctions, wang2020deep, Wang2021DeepPerspective}, which illustrates that deep learning enables researchers to extract economic information for practical uses. However, a particular challenge is how to design an effective operator to mix the numeric data and urban imagery, thus rendering this latent space predictive and informative.

This challenge is addressed by designing a mixing operator, which encodes the numeric and imagery data into a latent space with simple concatenation and supervised autoencoders. The latent variables are then imported into a simple behavioral predictor, which is similar to classical demand models. Section \ref{sec:review} reviews the literature about travel demand modeling, computer vision (CV), and urban computing applications. Section \ref{sec:theory} presents the design of the DHM framework with the mixing operators. Section \ref{sec:experiment} introduces data collection, local context, and experiment design. Section \ref{sec:results} investigates three empirical questions (1) whether the DHM framework can outperform demand models and deep learning models by effectively integrating the sociodemographics and satellite imagery (Section \ref{sec:results_performance}); (2) whether the latent space in DHM is spatially and socially meaningful (Section \ref{sec:results_latent_space}), and (3) DHM can generate new satellite imagery and derive the corresponding economic information for practical uses (Section \ref{sec:results_economic_interpretation}). Section \ref{sec:conclusion} summarizes our findings, limitations, and broad implications. To promote open science, our scripts have been uploaded to \url{https://github.com/sunnyqywang/demand_image/tree/DHM-AE}. 

\section{Literature review}
\label{sec:review}
\subsection{Travel demand modelling}

Demand models have been used extensively to analyze human decisions regarding travel mode choices \cite{Huan2021, wang2020deep,zheng2021equality}, adoption of new technologies \cite{Glerum2013, wang2019risk}, willingness to pay \cite{bansal2016assessing}, and behavioral loyalty \cite{Schmid2019,Irawan2020}. The most common demand models are the discrete choice models (DCM) based on the theory of random utility maximization. The data inputs of DCMs include individual- and alternative-specific attributes (e.g., sociodemographics, travel time, and cost), or psychometric features (e.g. perceptions and attitudes), which are often modeled using latent variables. Although the latent variables cannot be directly measured, they can be estimated with structural equation models \cite{Vij2016}. To integrate the latent variables and random utility theory, researchers developed hybrid demand models to jointly analyze the observed choice and latent variables, serving as a state-of-the-art demand model \cite{Walker2002, Ben-Akiva2002HybridChallenges}. The hybrid demand models have no theoretical limitations in the dimensions of the latent space; however, due to practical constraints such as model estimation algorithms, computational power, and the cost of data collection, the majority of the existing work adopted less than five latent variables without application to urban imagery \mbox{\cite{soto2018incentivizing, hess2018analysis, mahpour2018shopping, dada2019modelling, huan2022understanding}}. 

Increasingly, researchers started to enhance the demand models by accounting for complex input-output relationships with deep learning \cite{CANTARELLA2005121, lee2018comparison, Hillel2021, Cranenburgh2022, wu2020inductive, wu2021spatial, zhuang2020compound, zhuang2022uncertainty}. Deep neural networks (DNNs) take advantage of the increasingly available computation power and use gradient descent and regularization techniques (such as L1, L2, dropout, early stopping, and more) to enable the training of large networks with flexible architecture design. As a result, DNNs have shown superior performance in various applications \cite{Karlaftis2011, zheng2021equality}. However, DNNs are often criticized for the interpretability and instability problems \cite{paredes2017machine, hillel2020new}, which could be mitigated by gradient-based methods \cite{wang2020deep, van2021artificial, alwosheel2021did} or integrated DCM-DNN architectures \cite{wong2021reslogit, Wang2021, Sifringer2020}. The integrated architectures can facilitate researchers to derive economic information, enhance training efficiency, and stabilize the learning process. For example, the DNNs can fit the residuals of the DCM backbone \cite{wong2021reslogit, Wang2021, Sifringer2020}, learn decision rules \cite{van2019artificial}, and generate choice sets \cite{yao2022variational}. DNNs can also learn latent representations from survey indicators \cite{wong2018modelling}, personal characteristics and their interactions with the alternative attributes \cite{han2020neural}, GPS trajectories \cite{endo2016deep,yazdizadeh2021semi,yazdizadeh2019ensemble}, and embeddings for categorical variables that are typically hard to handle in traditional frameworks \cite{Arkoudi2021}. However, all existing studies use only numeric data, while deep learning is powerful because it can learn from high-dimensional data. There were some attempts at leveraging high-dimensional data, including treating GPS trajectories as images \cite{endo2016deep, yazdizadeh2021semi}, and stacking multiple features along a route to form 2D inputs \cite{yazdizadeh2019ensemble}, but none have leveraged real urban imagery. Real urban imagery is associated with various urban and socioeconomic characteristics, as shown by the vast number of studies in computer vision and urban computing.

\subsection{Computer vision and urban computing}
Rich urban and socioeconomic characteristics are contained in various urban images, including nightlight, satellite, and street view images. Using satellite imagery, researchers can learn land-use patterns \cite{Albert2017UsingScale}, quantify green cover \cite{Seiferling2017}, measure physical appearances of neighborhoods \cite{Naik2017}, and predict socioeconomic status \cite{Jean2016, Ayush2020, Yeh2020,Gebru2017}. In transportation, researchers used urban imagery to monitor traffic flows during the pandemic \cite{Chen2021}, predict pedestrian intentions using car and traffic cameras \cite{razali2021pedestrian}, and predict the usage of active modes based on street-view images \cite{hankey2021predicting}. Several studies also used street-view images to learn the perceived safety and general attitudes towards neighborhoods \cite{naik2014streetscore, dubey2016deep, zhang2018measuring}. Reviews of urban imagery applications can be found in \cite{Ibrahim2020UnderstandingAnalytics, Biljecki2021StreetReview}. However, we have not identified any study that used observed human decisions, such as travel behavior, as modeling outputs. 

Besides the predictive models, researchers are paying increasing attention to the generative models. As a baseline, the autoencoder (AE) can learn latent representations from images using an encoder and generate new images using a decoder. Even in its simplest form, the AE can learn latent representations, which are effective for prediction tasks \mbox{\cite{ren2018bearing, xu2020multi}}. A baseline AE can be enhanced by the multitask supervised learning as a regularization in its latent space, thus further improving its capacity of representation learning \mbox{\cite{dong2017autoencoder, huang2019supervised}}. However, AE is criticized for lacking sampling capacity and generating blurry images, so variational autoencoders (VAE) and generative adversarial networks (GAN) were proposed as two remedies. The VAE regularizes the latent space of AE with a Gaussian prior, enabling a smooth transition of generated images \mbox{\cite{ha2017neural}}. The GAN can generate realistic images through the game-theoretical training of its discriminator and generator. The loss terms in GAN can be augmented to the AEs to improve the manifold learning \mbox{\cite{makhzani2015adversarial, Dumoulin2017AdversariallyInference,xu2019adversarially}} and the quality of image generation \mbox{\cite{larsen2016autoencoding, Berthelot2019UnderstandingRegularizer, oring2020autoencoder}}. The state-of-the-art generative models integrate the loss functions of VAEs and GANs into the baseline AE, marking the convergence of the two lines of research \mbox{\cite{Esser_2021_CVPR,Rombach_2022_CVPR}}. Despite the proliferation of generative models in CV literature, no study has investigated how to generate satellite images from sociodemographic and travel characteristics.

More important than using travel behavior as another application, we have to design a novel approach drastically different from the existing urban computing research to model human decisions. Human decision is more complex than built environment labels (e.g. trees, parking lots, or other land use patterns), because it inevitably involves the discussions of social heterogeneity, economic implications, and human decision mechanisms. Unlike sociodemographic data, individual pixels in urban imagery do not have socioeconomic meanings, posing a unique challenge for deriving socioeconomic information from urban imagery. To address these challenges, we propose the DHM framework that encodes urban imagery and sociodemographics into a latent space, by which we could conduct associative analysis between sociodemographics, travel decisions, and satellite imagery. 

\section{Theory} \label{sec:theory}
\subsection{General framework of deep hybrid models}
The DHM can be represented as:
\begin{equation}
   y_n = g(z_n) = g(\mathcal{M}(x_n, I_n)) \\ 
   \label{eq:dhm_master}
\end{equation}

\noindent
in which $x_n$ and $I_n$ represent the numeric and imagery inputs, and $y_n$ represents the travel behavioral outputs. The two key components in DHM are  $\mathcal{M}(x_n, I_n)$, which combines the numeric sociodemographics $x_n$ and the urban imagery $I_n$, and $g(z_n)$, which predicts the travel outputs. In other words, the DHM framework consists of a \textit{mixing operator} $\mathcal{M(\cdot)}$ and a \textit{behavioral predictor} $g(\cdot)$. The behavioral predictor follows a generalized linear form: $g(z_n) = \sigma(\beta'z_n)$, in which $\sigma(\cdot)$ represents the link function and $\beta'z_n$ is a linear transformation of the latent variables $z_n = \mathcal{M}(x_n, I_n)$. In this formulation, $g(\cdot)$ is significantly simplified so that we could concentrate the discussion on the mixing operator. Meanwhile, $g(\cdot)$ is also flexible enough to accommodate a variety of output categories: single outputs, soft choice probabilities, and discrete choices. 

Table \ref{tab:mixing_operator_design} summarizes six models for the mixing operator $\mathcal{M}(x_n, I_n)$. We will provide an overview of the table in this section, followed by a detailed discussion of mixing operators in Section \ref{sec:supervision_mixing}. In Table \ref{tab:mixing_operator_design}, the first column explains the formula for computing the latent variables $z_n^{(i)}$. The second column presents the optimization formulation of neural network parameters. $E_\phi(\cdot)$ represents the encoder network parameterized by $\phi$, $D_\psi(\cdot)$ represents the decoder network parameterized by $\psi$, and $F_\omega(\cdot)$ represents the sociodemographic supervision network parameterized by $\omega$. The third column presents the latent dimensions of $z_n^{(i)}$, which vary drastically across the six models. 

\begin{table}[th!]
\caption{Design of the mixing operator}
    \centering
    \resizebox{0.9\linewidth}{!}{%
    \begin{tabular}{p{0.25\linewidth} | p{0.55\linewidth} | p{0.15\linewidth}}
        \toprule
        \textbf{Mixing operator} & \textbf{Optimization for parameter estimation} & \textbf{Latent dim} \\
        \midrule
        \multicolumn{3}{l}{\textbf{Panel 1: Benchmark Models}}\\
        \midrule
        \midrule
        \multicolumn{3}{l}{Model 1: Only sociodemographics (SD)} \\
        \midrule
        $z^{(1)}_n = x_n$ & N/A & 10  \\
        \midrule
        \multicolumn{3}{l}{Model 2: Only imagery with autoencoders (AE)} \\
        \midrule
        $z^{(2)}_n = E_{\hat{\phi}}(I_n)$ & $\hat{\phi}, \hat{\psi} = \underset{\phi, \psi}{\argmin} 
\; \mathcal{L}_{rec}$ & 18,432 \\
        \midrule
        \multicolumn{3}{l}{\textbf{Panel 2: Deep Hybrid Models}}\\
        \midrule \midrule
        \multicolumn{3}{l}{Model 3: Combining Models 1-2} \\
        \midrule
        $z^{(3)}_n = [z^{(1)}_n,z^{(2)}_n]$ & N/A & 10 + 18,432 \\ 
        \midrule
        \multicolumn{3}{l}{Model 4: Supervised autoencoder (SAE)} \\
        \midrule
        $z^{(4)}_n = E_{\hat{\phi}}(I_n)$ & $\hat{\phi}, \hat{\psi}, \hat{\omega} = \underset{\phi,\psi,\omega}{\argmin} \; \lambda \mathcal{L}_{rec} + (1-\lambda) \mathcal{L}_{sup}$ & 18,432 \\        
        \midrule    
        \multicolumn{3}{l}{Model 5: Supervised autoencoder with enhanced image regeneration (latent dim = 10)} \\
        \midrule
        $z^{(5)}_n = E_{\hat{\phi}}(I_n)$ & $\hat{\phi}, \hat{\psi}, \hat{\omega} = \underset{\phi,\psi,\omega}{\argmin} \; \lambda (\alpha_{rec} \mathcal{L}_{rec} + \alpha_{lpips} \mathcal{L}_{lpips} + \alpha_{GAN_G} \mathcal{L}_{GAN_G}  + \alpha_{KL} \mathcal{L}_{KL}) + (1-\lambda) \mathcal{L}_{sup}$ & 10 \\    
         \midrule    
        \multicolumn{3}{l}{Model 6: Supervised autoencoder with enhanced image regeneration (latent dim = 4,096)} \\
        \midrule
        $z^{(6)}_n = E_{\hat{\phi}}(I_n)$ & Same as Model 5 & 4,096 \\    
        \bottomrule
    \end{tabular}
    }
    \label{tab:mixing_operator_design}
\end{table}

Models 1 and 2 are benchmark models. Model 1 incorporates only sociodemographics $x_n$ without any imagery. Model 2 incorporates only the imagery information without including any sociodemographics by using the neurons of a baseline autoencoder as the latent variables. Autoencoders learn latent representations $z_n$ from image $I_n$ using an encoder $z_n=E_\phi(I_n)$, and reconstruct the image $\hat{I}_n$ using a decoder $\hat{I}_n=D_\psi(z_n)$, with $\phi$ and $\psi$ representing the parameters in the encoder and decoder. The basic autoencoder in Model 2 is trained with the reconstruction loss:
\begin{equation}
    \mathcal{L}_{rec}=\sum_n |I_n-\hat{I}_n|
\label{eq:rec}
\end{equation}
, which measures the L1 distance between the input and reconstructed images in the pixel space. Although the reconstruction loss is limited to only the pixels, other loss terms can be added to regularize the latent variables and improve the quality of representation learning (See Models 4-6).

\subsection{Mixing operator: supervised autoencoders} \label{sec:supervision_mixing}
Model 3 represents the simplest form of DHMs because it linearly concatenates the latent variables of Models 1 and 2. Although concatenation is simple, it is widely used to integrate diverse data sources in deep learning literature \mbox{\cite{kothari2021human, moreau2021data}}. By comparing the performances of Model 3 to Models 1-2, we could observe whether the two data sources are complementary. This simple concatenation in Model 3 serves as a benchmark of all the DHMs.

\begin{figure}[ht!]
    \centering
    \includegraphics[width=0.6\linewidth]{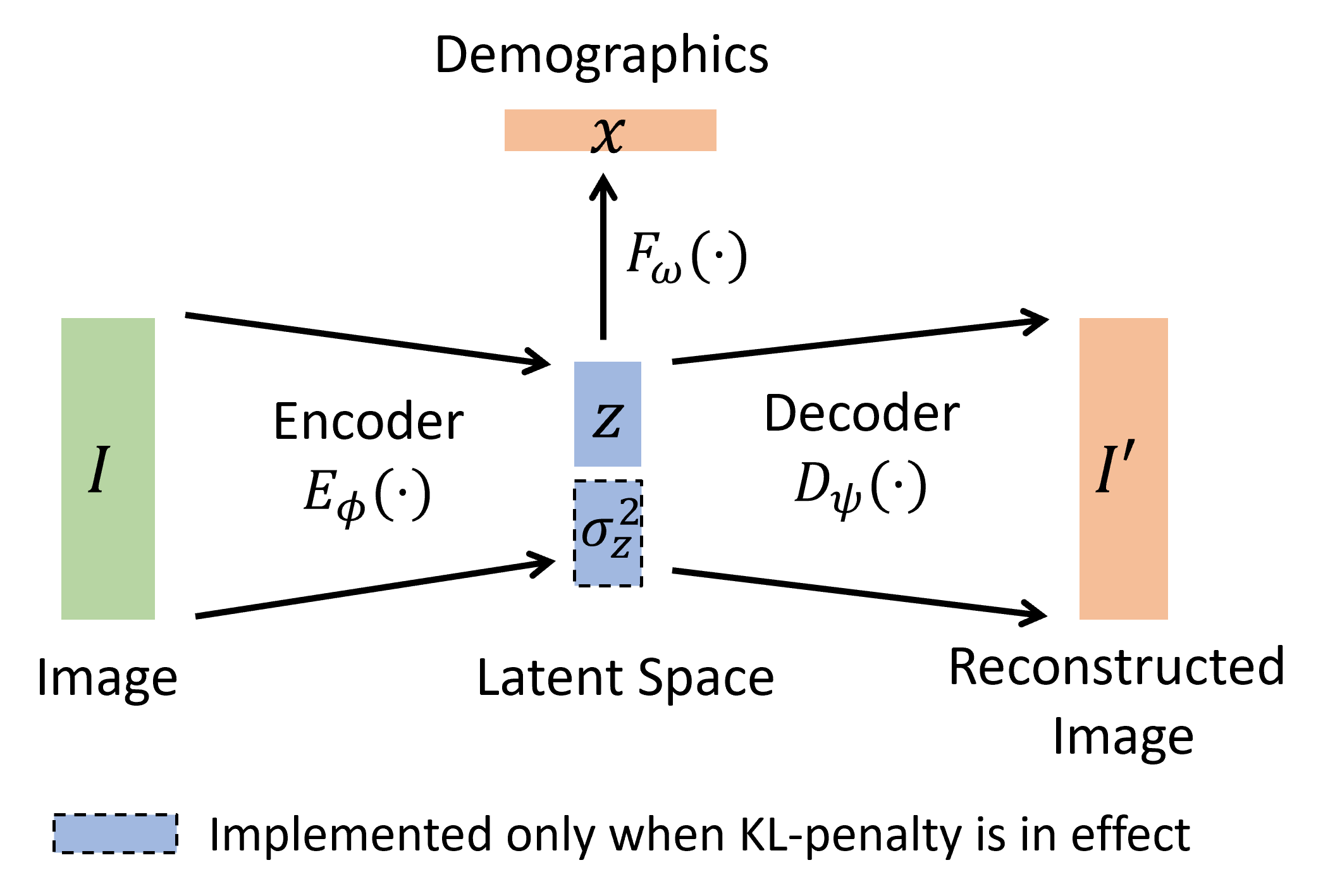}
    \caption{Supervised Autoencoders}
    \label{fig:supervision_as_mixing}
\end{figure}

Models 4-6 are designed as the supervised autoencoders (SAE), the diagram of which is visualized in Figure \ref{fig:supervision_as_mixing}. Different from a baseline AE, the latent variables $z_n$ of SAEs can reconstruct images through the decoder $D_\psi(z_n)$ and sociodemographics through a supervision network $F_\omega(z_n)$. The supervision of sociodemographics enhances the stability in the latent space because it extends the basic AE to multitask learning with the sociodemographic supervision loss as regularization \cite{Le2018SupervisedRegularizers}. Both multitask learning and regularization can constrain the latent space and improve training stability \cite{Caruana1996PromotingOutputs, Maurer2016TheLearning}. 

The SAEs (Models 4-6) are trained by minimizing the sum of an AE loss $\mathcal{L}_{AE}$ and a sociodemographic supervision loss $\mathcal{L}_{sup}$. 
\begin{equation}
\mathcal{L}_{SAE} = (1-\lambda) \mathcal{L}_{AE} + \lambda \mathcal{L}_{sup} \\
\label{eq:sae}
\end{equation}
in which $\mathcal{L}_{sup}$ represents the absolute error of the sociodemographic variables $x$ weighted by an adaptive hyperparameter $\alpha_{sup}$:
\begin{equation}
    \mathcal{L}_{sup} = \alpha_{sup} |x - \hat{x}|
    \label{eq:sup}
\end{equation}
\noindent In Equation \ref{eq:sae}, a mixing hyperparameter $\lambda \in [0,1]$ measures the trade-off between image reconstruction and sociodemographic supervision. When $\lambda = 0$, the latent space contains only imagery information. When $\lambda = 1$, the latent space contains only sociodemographic information. Equation \ref{eq:sae} represents the most general form of the loss function shared across Models 4-6, but the AE loss $\mathcal{L}_{AE}$ varies between Model 4 and Models 5-6. 

Model 4 is a baseline SAE using the L1 reconstruction loss (Equation \ref{eq:rec}) as the $\mathcal{L}_{AE}$. Therefore, the loss function of Model 4 is:
\begin{equation}
\mathcal{L}_{SAE} = (1-\lambda) \mathcal{L}_{rec} + \lambda \mathcal{L}_{sup} \\
\label{eq:model_4}
\end{equation}

Although Models 5-6 share the same general diagram (Figure \ref{fig:supervision_as_mixing}) and training objective (Equation \ref{eq:sae}) as Model 4, they significantly expand upon the AE loss $\mathcal{L}_{AE}$ into a list of loss terms, which can effectively enhance the quality of image generation. The AE loss in Models 5-6 includes the reconstruction loss $\mathcal{L}_{rec}$, the KL divergence loss between the estimated latent space distribution and the standard normal prior distribution $\mathcal{L}_{KL}$,  the learned perceptual image patch similarity (LPIPS) $\mathcal{L}_{lpips}$, and the adversarial objective $\mathcal{L}_{GAN_G}$ and $\mathcal{L}_{GAN_D}$\cite{Rombach_2022_CVPR}. The KL-penalty is introduced in the same way as VAE to enable smooth transitions and sampling in the latent space, and constrain the magnitude of the variance in the latent space. The LPIPS loss and the adversarial objective enhance the realism of the generated images and mitigate the blurriness caused by relying solely on pixel-space metrics such as the L1 reconstruction loss.The full breakdown is shown in Equation \ref{eq:ae+}: 
\begin{equation}
\mathcal{L}_{AE} = \alpha_{rec} \mathcal{L}_{rec}  + \alpha_{KL} \mathcal{L}_{KL} + \alpha_{lpips} \mathcal{L}_{lpips} + \alpha_{GAN_G} \mathcal{L}_{GAN_G}\\
\label{eq:ae+}
\end{equation}
\noindent in which all the $\alpha$ terms are the adaptive hyperparameters, which will be introduced later. Both Models 5 and 6 use Equation \ref{eq:ae+} as the objective function, although they are designed with different latent dimensions (10 vs. 4,096) to test the impacts of latent dimensions on prediction and image generation. In Equation \ref{eq:ae+}, The first loss term is still the reconstruction loss $\mathcal{L}_{rec}$ as Equation \ref{eq:rec} - the pixel level L1 distance between input and reconstructed images. 

The KL divergence $\mathcal{L}_{KL}$ measures the distance between the calculated and the standard Gaussian distributions, as shown in Equation \ref{eq:kl}. The latent variables use $k$-dimensional diagonal Gaussian distribution with estimated mean $\hat{z}_n$ and variance $\hat{\sigma}^2_{z_n}$.
\begin{equation}
    \mathcal{L}_{KL} = \sum_n KL(\mathcal{N}(\hat{z}_n,\hat{\sigma}^2_{z_n}) || \mathcal{N}(0,1)) = \frac{1}{2} \sum_n \sum_k ( \hat{z}_{n,k}^2 + \hat{\sigma}^2_{z_{n,k}} - 1 - log ( \hat{\sigma}^2_{z_{n,k}}) )
    \label{eq:kl}
\end{equation}

The $\mathcal{L}_{lpips}$ loss - learned perceptual image patch similarity (LPIPS), also known as perceptual loss $\mathcal{L}_{lpips}$ - is widely used to enhance perceptual similarity between the deep features of two images \cite{zhang2018the}. To calculate the LPIPS loss between the original and reconstructed images $I_n$ and $\hat{I}_n$, the feature stacks from $L$ layers are extracted from a pre-trained neural network, and unit-normalized in the channel dimension, which is designated as $\hat{y}^l_{I_n, hw}, \hat{y}^l_{\hat{I}_n, hw} \in \mathbf{R}^{H_l \times W_l \times C_l}$. The features are then scaled by vector $w_l$ before computing the L2 distance. Its formula is:

\begin{equation}
    \mathcal{L}_{lpips} = \sum_n \sum_l \frac{1}{H_lW_l} \sum_{h,w} ||w_l \odot (\hat{y}^l_{I_n,hw} - \hat{y}^l_{\hat{I_n}, hw}) ||^2_2 
    \label{eq:lpips}
\end{equation}

Lastly, the generative adversarial loss uses a discriminator network $Disc(\cdot)$ to differentiate between real and generated images. Equation \ref{eq:gan_g} shows the generator (decoder $D_\psi$) loss function $\mathcal{L}_{GAN_G}$, and Equation \ref{eq:gan_d} shows the discriminator loss function $\mathcal{L}_{GAN_D}$. GANs are trained in a game-theoretic manner with alternating generator and discriminator updates. At each step, the model first trains the generator (decoder), the encoder, and the supervision network while fixing the discriminator network weights. It then trains the discriminator network with the other models fixed.
\begin{equation}
    \mathcal{L}_{GAN_G} = \sum_n{ - Disc(\hat{I_n})}
    \label{eq:gan_g}
\end{equation}
\begin{equation}
    \mathcal{L}_{GAN_D} = \sum_n{max(0,1+Disc(I_n)) + max(0, 1-Disc(\hat{I_n}))}
    \label{eq:gan_d}
\end{equation}

The loss terms are associated with adaptive hyperparameters, including $\alpha_{sup}$, $\alpha_{rec}$, $\alpha_{KL}$, $\alpha_{lpips}$, and $\alpha_{GAN_G}$. These adaptive hyperparameters dynamically balance the loss terms and stabilize the complex training process. For example, $\alpha_{sup}$ is calculated by the ratio of the loss terms' gradients with respect to the parameter magnitude in the last layer $L$ of the decoder $D_\psi$ \cite{Esser_2021_CVPR} using $\delta=10^{-6}$ for numerical stability. 
\begin{equation}
    \alpha_{sup} = \frac{\nabla_{G_L} [\mathcal{L}_{rec}]}{\nabla_{G_L} [\mathcal{L}_{sup}]+\delta}
    \label{eq:l_sup}
\end{equation}
\noindent The $\alpha_{GAN_G}$ is also adaptive and calculated similarly as $\alpha_{sup}$. 
\begin{equation}
    \alpha_{GAN_G} = \frac{\nabla_{G_L} [\mathcal{L}_{rec}]}{\nabla_{G_L} [\mathcal{L}_{GAN_G}]+\delta}
    \label{eq:l_gan}
\end{equation}
To simplify the training process, other adaptive hyperparameters are fixed using the default setting from the stable diffusion models\footnote{\url{https://github.com/CompVis/stable-diffusion/blob/main/configs/autoencoder/autoencoder_kl_8x8x64.yaml}}, where $\alpha_{rec}=1, \alpha_{KL}=1e^{-6}$, and $\alpha_{lpips}=1$. The adaptive hyperparameters $\alpha$'s and the static hyperparameter $\lambda$ serve different purposes: the former mainly stabilizes the training process and the latter diagnoses the trade-off between image reconstruction and sociodemographic supervision. 




With the SAE design, Models 4-6 interact sociodemographics and satellite imagery through the latent space. Although the formula $E_{\hat{\phi}}(I_n)$ contains only images as inputs, the encoder parameters $\hat{\phi}$ are trained using both $x_n$ and $I_n$. Therefore, the $\hat{\phi}$ absorbs the information from two data structures, so it enables the latent variables $E_{\hat{\phi}}(I_n)$ to blend information from sociodemographics and urban imagery. The design of the SAE models is highlighted due to its similarity to the classical hybrid demand model structure \cite{Ben-Akiva2002HybridChallenges}. In classical hybrid demand models, a structural component describes how the sociodemographic variables relate to travel behavior, and a measurement model describes the relationship between observed variables and latent variables \cite{Walker2002, Ben-Akiva2002HybridChallenges}. Our DHMs resemble this structural approach mainly through the multitask learning framework, and we further enrich the existing hybrid models by leveraging the higher-dimensional deep architecture to design the latent space and incorporate urban imagery. 

\subsection{Behavioral predictor}
\label{sec:methods_behavior_predictor}
The behavioral predictor $g(z)$ is designed as a generalized linear regression: 

\begin{equation}
    \hat{y} = g(z) = \sigma(\beta'z)
\label{eq:behavioral_predictor}
\end{equation}

\noindent Despite its simplicity, Equation \ref{eq:behavioral_predictor} is sufficiently flexible to accommodate three output variables: (1) aggregate travel mode shares as individual outputs, (2) aggregate travel mode shares as a joint output, and (3) individual travel mode choices. A general form of training the behavioral predictor is:

\begin{equation}
    \underset{\beta}{\argmin} \ \mathcal{L}(y_n, \sigma(\beta'z_n)) + \theta||\beta||_p
\label{eq:general_training}
\end{equation}

\noindent in which $\mathcal{L}(y_n, \sigma(\beta'z_n))$ represents the loss function (e.g. mean squared error or negative log-likelihood), $\beta$ represents the parameters, $\theta$ is the sparsity hyperparameter to control the sparsity of $\beta$. In all three examples, the sparsity hyperparameter $\theta$ serves as the weight for L1 (LASSO) regularization to address overfitting and to identify the relevant latent variables. The LASSO approach is essential in reducing the potential overfitting from the high-dimensional latent space in Models 2, 3, 4, and 6 in Table \ref{tab:mixing_operator_design}. The three outputs are clarified in the following three subsections.

\subsubsection{Aggregate travel mode shares as separate outputs}
The first example is relatively straightforward: since the outcome $y_n$ is a continuous scalar to represent the aggregate travel mode shares, the link function $\sigma$ is designed as an identity mapping: 
\begin{equation}
\sigma(\beta'z_n) = \beta'z_n
\label{eq:agg1}
\end{equation}
The training uses the mean squared error as the objective. 

\subsubsection{Aggregate travel mode shares as a joint output} \label{sec:agg_model}
The second example uses the aggregate travel mode shares as a joint output of the model. Since the joint mode shares should add up to one, a softmax function is specified as the link function to implement this constraint. Using the linear transformation on $\beta'z_n$, we could represent the output mode shares as: 
\begin{equation}
    \begin{split}
        P_{nk} =  \frac{e^{V_{nk}}}{ \sum_j e^{V_{nj}}} = \frac{e^{\beta'_k z_n}}{ \sum_{j} e^{\beta'_{j} z_n}}
    \end{split}
    \label{eq:agg2}
\end{equation}

\noindent in which $P_{nk}$ represents the market shares of mode $k$ in region $n$. Since the market shares follow a probability distribution, Kullback-Leibler (KL) loss is used to specify the general loss in Equation \ref{eq:general_training}:
\begin{equation}
\mathcal{L}_{KL} = \frac{1}{N} \sum_{n=1}^{N}{\sum_{k=1}^{K}{P_{nk}(\ln{P_{nk}} - \ln{\hat{P}_{nk}})}}
\end{equation}

\subsubsection{Disaggregate travel mode choice}
The third example uses individual travel mode choice as the model outputs. The choice probabilities of trip $n$ for alternative $k$ are specified as 
\begin{equation}
P_{nk} = \frac{e^{V_{nk}}}{ \sum_j e^{V_{nj}}}
\end{equation}
\noindent where $P_{nk}$ is the probability of trip $n$ taken with alternative $k$, $V_{nk}$ is the utility of alternative $k$ for trip $n$. Since the disaggregate travel mode choice involves origin-destination pairs, the alternatives' attributes for each OD pair concatenate the origin $o_n$ and destination $d_n$: $z_n = [z_{o_n}, z_{d_n}]$. Similar to the aggregate travel mode analysis, the utility function takes a simple linear form as $V_{nk} = \beta^{'}_{k} z_n$. However, since discrete choice models have unique data structures in the input variables, which consist of individual- and alternative-specific variables, the detailed specification is slightly different from the aggregate travel mode analysis (Appendix I). Let $y_{nk}$ represent the observed mode ($y_{nk}=1$ if mode $k$ is used for trip $n$, and $y_{nk}=0$ otherwise), and $N$ is the total number of trips. The training loss in Equation \ref{eq:general_training} is substantiated by the cross entropy loss:
\begin{equation}
   \mathcal{L}_{ce} = \frac{1}{N} \sum_{t=1}^{N}{\sum_{k=1}^{K}{-y_{nk}\ln{P_{nk}}}}
\end{equation}


\subsection{Deriving economic information from generated satellite imagery}
\label{sec:methods_econ_info}
Recent work has demonstrated that deep learning models can provide economic information as complete as the classical discrete choice models\cite{wang2020deep}. Building upon deep learning, the DHMs can empower researchers to compute economic information from generated satellite images. Here we provide the formula for computing market shares, substitution patterns of alternatives, social welfare, and choice probability derivatives with highlights on the latent variables $z$ that connect images, sociodemographics, and travel behaviors. Specifically, the market share of alternative $k$ can be computed as  
\begin{equation}
    s_{k} = \sum_n {P}_{nk} = \sum_n \frac{e^{V_{nk}}}{ \sum_j e^{V_{nj}}} 
    \label{eq:market_share}
\end{equation} 
\noindent Social welfare of individual $n$ takes the standard logsum formula: 
\begin{equation}
\label{eq:welfare}
    \frac{1}{\alpha_n}  \log (\sum_{j} e^{{V}_{nj}}) 
\end{equation}
where $\alpha_n$ measures the marginal utility of income that translates social welfare into dollar values. The substitution pattern for two alternatives $j$ and $k$ is defined as the ratio of their choice probabilities: 
\begin{equation}
    \frac{P_{nj}}{P_{nk}} = \frac{e^{V_{nj}} / \sum_{k'}e^{V_{nk'}} }{e^{V_{nk}} / \sum_{k'}e^{V_{nk'}}} = e^{V_{nj}-V_{nk}}
    \label{eq:sub_pattern}
\end{equation}

Although the three equations above appear similar to those in the standard demand models, they already incorporate satellite imagery to compute the utility value $V_{nk}$ through the latent variable $z$. Among the six models in Table \mbox{\ref{tab:mixing_operator_design}}, the DHMs (Models 3-6) embed both satellite images and sociodemographic variables into the latent variable $z_n$. Although only images are used as inputs in Models 4-6, the parameters $\hat{\phi}$ are the functions of both sociodemographics and images, thus encoding their information into the latent variables $E_{\hat{\phi}}(I_n)$. Any latent variable $\tilde{z}$ can be used to visualize satellite imagery through the decoder as $\tilde{I}=D_\psi(\tilde{z})$, and compute the utility, market shares, social welfare, and substitution patterns with Equations \ref{eq:market_share}, \ref{eq:welfare}, and \ref{eq:sub_pattern}.

Using the DHMs, we could further compute a directional gradient of choice probabilities regarding the latent variable $z$:
\begin{equation}
    \nabla_u P_{nk}({z}) = {u} \cdot \nabla P_{nk}({z}) 
    \label{eq:prob_derv}
\end{equation}
where $\nabla P_{nk}({z})$ represents the gradient of choice probability regarding the latent variable $z$, and ${u}$ represents a direction to move in the latent space. Since DHMs connect $z$ and urban imagery $I$, this directional gradient can describe the sensitivity of choice probabilities with respect to the satellite image $I_n$ corresponding to the latent variable $z_n$. For example, the direction can be defined as one-directional as ${u}={z_2}-{z_1}=E(I_2)-E(I_1)$ by using two existing images $I_1$ and $I_2$, or extended to a multi-directional movement by linearly combining multiple $u$'s: ${u} = a_1{u_1}+a_2{u_2}$, where $u$ takes into account two directions ${u_1}$ and ${u_2}$. In our empirical analysis, we will demonstrate the directional sensitivity of choice probabilities regarding satellite images by discretizing the latent space. For example, we can create a directional vector $u = z_2 - z_1$ to define the directional movement, and compute the choice probabilities and satellite images using a new latent variable $\tilde{z} = {z_1} + a{u} = {z_1} + a({z_2} - {z_1})$, in which $a$ ranges between zero and one. With this approach, we can visualize a new satellite image $\tilde{I}$ through the decoder as $\tilde{I} = D_\psi(\tilde{z})$ and compute the choice probabilities through the behavioral predictor as $g(\tilde{z})$. We can also compute market shares, social welfare, and substitution patterns by applying $V_{nk}(\tilde{z})$ to Equations \ref{eq:market_share},  \ref{eq:welfare}, and \ref{eq:sub_pattern}. 

\section{Experiment design}
\label{sec:experiment}
\subsection{Data}
\begin{figure}[ht!]
    \centering
    \includegraphics[width=\linewidth]{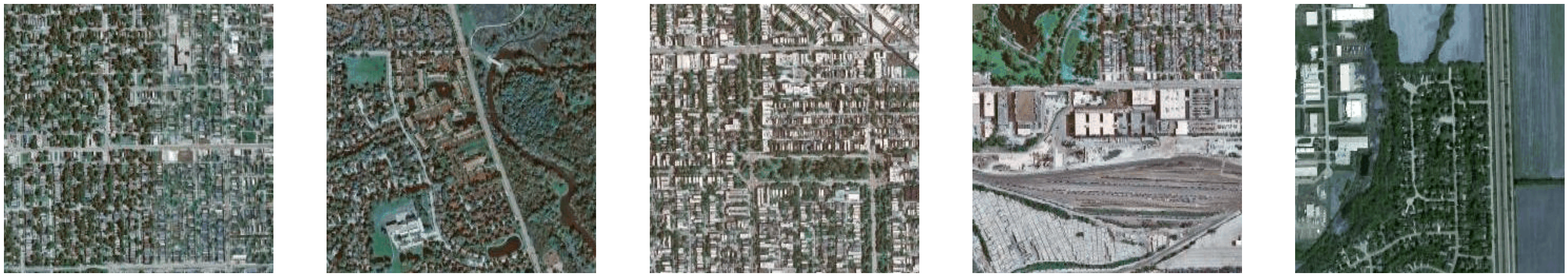}
    \caption{Samples of satellite images}
    \label{fig:image_sample}
\end{figure}

The experiments combine satellite images from Google API, socio-demographics from the census, and individual travel behavior from MyDailyTravel Chicago Survey in 2018-2019 \cite{MyHub}. The same number of satellite images were sampled from each census tract, which are then linearly combined in the latent space of the DHM for each region $s$ by $z_s = \frac{1}{I_s} \sum_{i=1}^{I_s}{z_i}$, where $z_s$ is the latent representation averaged over $I_s$ satellite images. The sociodemographics are obtained from the American Community Survey (ACS)\footnote{\url{https://www.census.gov/programs-surveys/acs/data.html}}, which includes total population, age groups, racial composition, education, economic status, and travel information (e.g. commuting time). After data preprocessing, The dataset has $1,571$ census tracts with $80$K observed trips. Figure \ref{fig:image_sample} shows five samples of the standard satellite images with a 256 $\times$ 256 size. 

The MyDailyTravel survey provides individual sociodemographics and trip attributes for the disaggregate analysis. It contains $42$ variables, including time of the trip (morning/after), trip distance, home-based trips, trip purposes, perceived time importance, age, disability, and education, among many other travel and social factors. The initial travel modes have more than ten categories, but they are aggregated into auto, active (walk+bike), public transit, and others, thus creating a relatively balanced choice set. 

\subsection{Model training}
\begin{figure}[ht!]
    \centering
    \includegraphics[width=\linewidth]{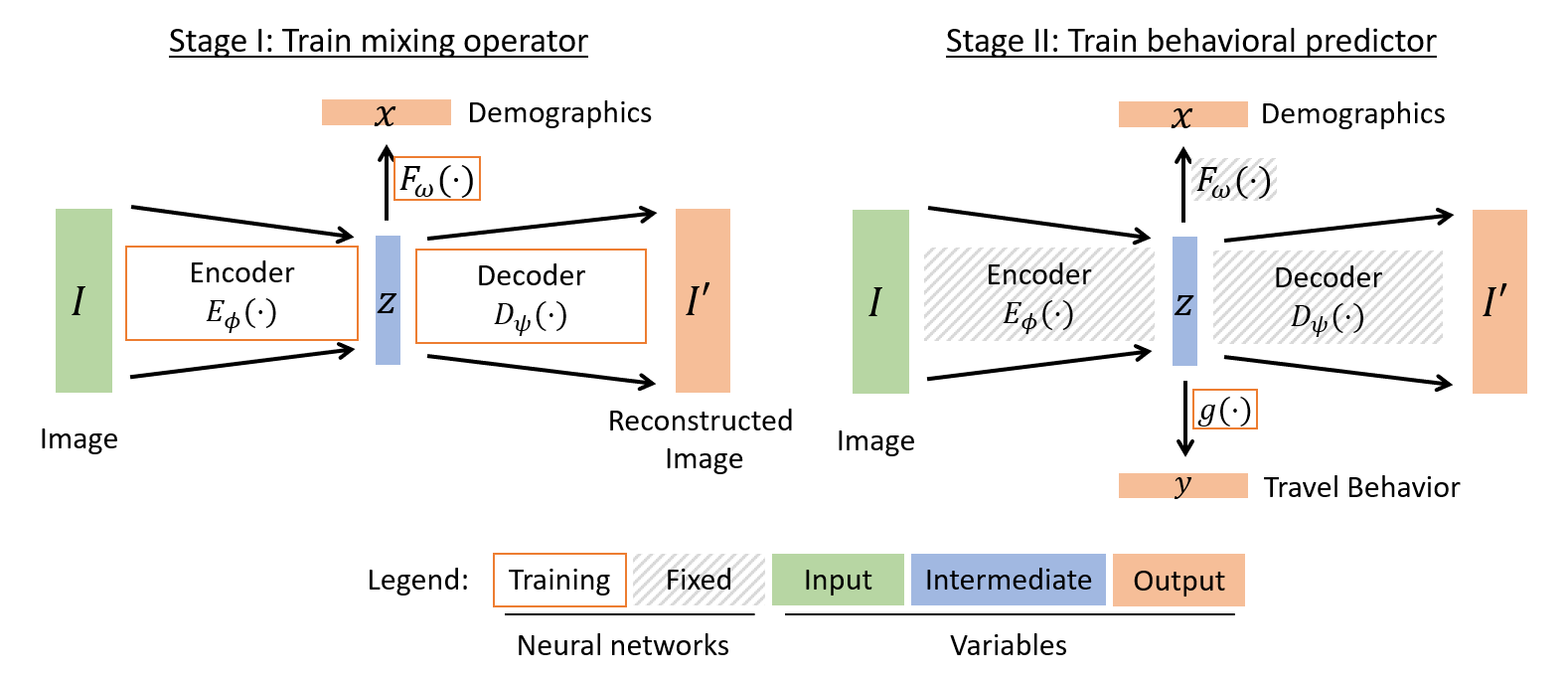}
    \caption{Two-step training}
    \label{fig:training}
\end{figure}

The training of DHMs takes two steps - training the mixing operator and then the behavioral predictor - as summarized in Figure \ref{fig:training}. During stage I, we train the SAEs, which consists of the encoder $E_\phi(I)$, decoder $D_\psi(z)$, and the sociodemographic predictor $F_\omega(z)$. Every input image is encoded into a latent variable $z$, which then passes through $F_\omega(z)$ and $D_\psi(z)$ to reconstruct sociodemographics and images. During stage II, only the behavioral predictor $g(z)$ is trained. In testing, an input image is encoded into a latent variable $z$, which then predicts sociodemographics and travel behaviors through $F_\omega(z)$ and $g(z)$, and generates images through $D_\psi(z)$. 

The mixing hyperparameter $\lambda$ is searched on a linear scale ($\lambda = 0.1,0.3,0.5,0.7,0.9$) to evaluate the impacts of the mixing on model performance. The sparsity hyperparameter $\theta$ is searched through all positive values to evaluate the sparsity effects. In model evaluation, the five-fold cross validation is used. The $R^2$ is calculated on the auto, active, and public transit separately to evaluate the aggregate models. The cross-entropy loss and accuracy rates are used to evaluate the disaggregate models. 


The six models require drastically different levels of computational resources. Models 2-4 are trained on an NVIDIA 2080 Ti GPU with 11GB RAM for the 52 million parameters in the ResNet-50 architecture \mbox{\cite{He2016DeepRecognition}}. The Models 5-6 are trained on a NVIDIA V100 GPU with 32GB RAM, which costs around \$15,000 at the current market price, for the 126 million parameters in the U-Net from the stable diffusion models. Besides monetary costs, the relatively simple model (e.g. Model 4) takes around 12 hours to train, and the more complex ones (e.g. Model 6) take two to three days. Such a significant amount of computational resources could inhibit the academic communities from replicating this study. Therefore, we present all six models with distinctly different degrees of complexity, thus facilitating the replication of at least Models 1-4 for the researchers with limited computational resources. In fact, the advanced Model 6 does not guarantee a better result on all the fronts. As shown below, it dominates others by the quality of image regeneration but generates predictive performance nearly equivalent to that of Model 4.

\section{Results}
\label{sec:results}

\subsection{Predictive performance}
\label{sec:results_performance}
Table \ref{tab:performance_table} summarizes the empirical results of the six models in three travel behavior tasks. Panels 1-2 present the aggregate travel demand analysis using travel modes as separate and joint outputs, and Panel 3 the disaggregate travel demand analysis. The aggregate models (Panels 1 and 2) typically use ten sociodemographic variables and 18,432- or 4,096-dimensional latent variables to represent urban imagery. The disaggregate analysis (Panel 3) incorporates an additional 42 travel attributes into the input variables. The six columns correspond to the six models in Table \ref{tab:mixing_operator_design}. The first row in each panel illustrates the dimension of latent variables. Each entry reports the training/testing performance, selected by the highest testing performance with the optimal $\lambda$ and $\theta$ values using five-fold cross-validation. The hyperparameter selection with other $\lambda$ and $\theta$ values can be found in Appendix II. Because of its superior quality in image regeneration, only the Model 6 with $\lambda=0.7$ is used for the further analysis in Sections \ref{sec:results_latent_space} and \ref{sec:results_economic_interpretation}. The predictive performance yields five major findings as follows.


\begin{table}[t!]
    \centering
    \caption{Predictive performance}
    \resizebox{\linewidth}{!}{
    \begin{tabular}{l|c|c|c|c|c|c}
        \toprule
        Model & 1 & 2 & 3 & 4 & 5 & 6 \tabularnewline
         & SD & AE & [SD$|$AE] & Basic SAE & SAE (10) & SAE (4096) \tabularnewline
        \midrule
        \multicolumn{7}{l}{\textit{Panel 1: Aggregate Mode Choice as Separate Outputs - Linear Regression}}\tabularnewline
        \midrule
        Latent Dim & 10 & 18432 & 10+18432 & 18432 & 10 & 4096 \tabularnewline
        \midrule
        Auto ($R^2$) & 0.553/0.541 & 0.627/0.532 & 0.655/0.594 & \textbf{0.720/0.644} & 0.579/0.566 & 0.676/0.640 \tabularnewline
        Active ($R^2$) & 0.456/0.446 & 0.542/0.407 & 0.463/0.472 & \textbf{0.583/0.531} & 0.472/0.457 & 0.566/0.524 \tabularnewline
        PT ($R^2$) & 0.443/0.424 & 0.503/0.395 & 0.543/0.464 & 0.545/0.482 & 0.462/0.443 & \textbf{0.520/0.483}\tabularnewline
        \midrule 
        \multicolumn{7}{l}{\textit{Panel 2: Aggregate Mode Choice as a Joint Output - Multinomial Regression}}\tabularnewline
        \midrule
        Latent Dim & 10 & 18432 & 10+18432 & 18432 & 10 & 4096 \tabularnewline
        \midrule
        KL Loss & 0.146/0.149 & 0.149/0.158 & 0.135/0.146 & \textbf{0.111/0.128} & 0.142/0.144 & 0.120/0.129 \tabularnewline
        Auto ($R^2$) & 0.544/0.535 & 0.514/0.484 & 0.596/0.557 & \textbf{0.704/0.643} & 0.577/0.568 & 0.669/0.637 \tabularnewline
        Active ($R^2$) & 0.448/0.439 & 0.430/0.396 & 0.500/0.452 & \textbf{0.627/0.524} & 0.471/0.453 & 0.566/0.515 \tabularnewline
        PT ($R^2$) & 0.424/0.407 & 0.396/0.364 & 0.476/0.428 & 0.569/0.476 & 0.436/0.427 & \textbf{0.543/0.496} \tabularnewline
        \midrule 
        \multicolumn{7}{l}{\textit{Panel 3: Disaggregate Mode Choice - Discrete Choice Analysis}} \tabularnewline
        \midrule
        Latent Dim & 42+10*2 & 18432*2 & 42+10*2+18432*2 & 42+18432*2 & 42+10*2 & 42+4096*2 \tabularnewline
        \midrule
        CE Loss & 0.423/0.422 & 0.659/0.652 & \textbf{0.380/0.389} & 0.378/0.409 & 0.405/0.415 & 0.373/0.404\tabularnewline
        Accuracy & 0.856/0.857 & 0.756/0.759 & \textbf{0.871/0.868} & 0.868/0.855 & 0.862/0.859 & 0.871/0.856\tabularnewline
        \midrule
        \multicolumn{7}{l}{Note: each entry is represented as training/testing performance.} \tabularnewline
        \bottomrule
    \end{tabular}
    }
    \label{tab:performance_table}
\end{table}

First, both the sociodemographics and satellite imagery are informative for travel behavioral analysis because the predictive performance of Models 1-2 is substantially higher than zero across all three panels. In Panel 1, Model 2 using only satellite imagery can explain 53.2\%, 40.7\%, and 39.5\% of the variations in automobiles, active travel modes, and public transit (Column 2), slightly lower than 54.1\%, 44.6\%, and 42.4\% in Model 1. This finding still holds in Panel 2, while Model 2 performs much worse than Model 1 in the disaggregate analysis (Panel 3). The results are quite reasonable: urban imagery is slightly less informative than the sociodemographics in predicting aggregate travel behavior, while it is much less so than sociodemographics and particularly travel attributes, which are the major incentives determining individual travel behavior. Regardless of these variations, the performance of Models 1-2 demonstrates evidently that both sociodemographics and satellite imagery can assist in travel behavioral prediction. 

\begin{figure}[ht!]
    \centering
    \resizebox{0.9\linewidth}{!}{
    \includegraphics{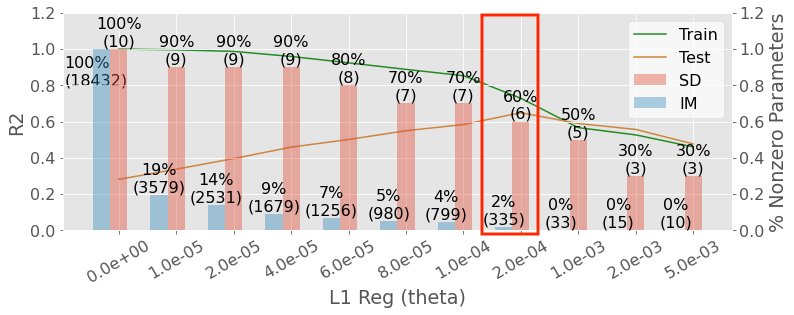}}
    \caption{Non-zero coefficients from sociodemographics and urban imagery in Model 3 with various $\theta$ values.}
    \label{fig:comp_plot}
\end{figure}

Second, the numeric data and satellite imagery are complementary, as shown by the performance comparison in Table \ref{tab:performance_table} and the information decomposition in Figure \ref{fig:comp_plot}. Across all three panels, Models 3-6 consistently outperform Models 1-2, indicating that it is more effective by mixing rather than independently using the two data structures. Since Model 3 simply concatenates the latent variables of Models 1-2, its higher performance is attributed to data complementarity rather than our DHM design, which exists only in Models 4-6. Besides comparing performance, the non-zero coefficients in Model 3 also indicate data complementarity. Figure \ref{fig:comp_plot} illustrates how the non-zero coefficients decrease with a larger sparsity hyperparameter $\theta$. With the optimal $\theta=2e^{-4}$, the non-zero coefficients origin from both the sociodemographic and urban imagery sides, consisting of six sociodemographic variables and 335 imagery variables, demonstrating the complementary contributions to prediction from the two data structures. 

Third, the SAEs are more effective than the simple concatenation in the aggregate contexts (Panels 1-2) but less so in the disaggregate contexts. Model 4 outperforms Model 3 by 5-10\% in $R^2$ and about 12\% in the KL loss in Panels 1-2. This result suggests that the nonlinear mixing of sociodemographics and urban imagery through the SAEs is more effective than a linear concatenation in the aggregate analysis. Across the three travel modes, the performance improvement is more significant in the automobile and active modes, suggesting that the built environment represented by satellite imagery is more informative for the usage of private vehicles, walking, and cycling than public transit. The effects of SAEs appear less evident in the disaggregate analysis because the linear concatenation in Model 3 outperforms Models 4-6, suggesting that the travel attributes influence individual travel choices in a linear way. 

Fourth, it is important to create a relatively high-dimensional latent space (Model 6) for high predictive performance, although the low-dimensional latent space (Model 5) can also explain substantial variation in travel behaviors. Models 5-6 were designed to be identical in data, model, and training but differed in only the latent dimensions (10 vs. 4,096). Across the three panels, Model 6 consistently outperforms Model 5 in predictive performance, suggesting the importance of a relatively high-dimensional latent space for effective prediction. However, Model 5 also achieves substantial predictive performance, reaching at least 80-90\% of the predictive performance in Model 6. This finding suggests that the explanatory power in the latent variables has a somewhat long-tail distribution: it is concentrated in the first few principal dimensions but also spread into hundreds of others.

\begin{figure}[ht!]
    \centering
    \includegraphics[width=0.8\linewidth]{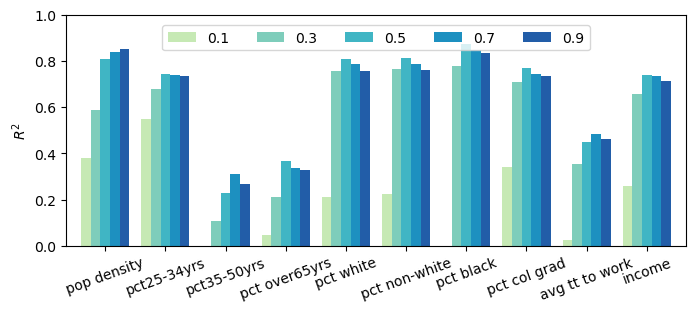}
    \caption{Performance of predicting sociodemographics in Model 6 with various $\lambda$ values}
    \label{fig:lambda_effect}
\end{figure}

Lastly, the DHMs can achieve relatively high performance for predicting the sociodemographic variables when the $\lambda$ value gradually deviates from zero. Figure \ref{fig:lambda_effect} visualizes how the $R^2$ in predicting sociodemographics varies with $\lambda$ for every sociodemographic variable. Since $\lambda$ balances the image reconstruction and the sociodemographic prediction, a higher $\lambda$ value can instill more sociodemographic information into the latent space. As shown in Figure \ref{fig:lambda_effect}, a relatively high $\lambda$ value can significantly improve the predictive performance of the sociodemographics, including population density, age, racial shares, and income. With an optimal $\lambda$ value at around 0.5-0.7, the predictive performance achieves a relatively high $R^2$ (around 80\%) in population density, racial structures, education, and income, but relatively lower $R^2$ (around 30-50\%) in age and average travel time. 

\subsection{Navigating the latent space}
\label{sec:results_latent_space}
With the design of DHMs, the latent space encodes the sociodemographic and urban imagery information. The latent space can be understood by using dimension reduction techniques, including K-Means to identify the discrete latent clusters and t-distributed stochastic neighbor embedding (tSNE) for continuous embedding. 

\begin{figure}[ht!]
    \centering
    \includegraphics[width=\linewidth]{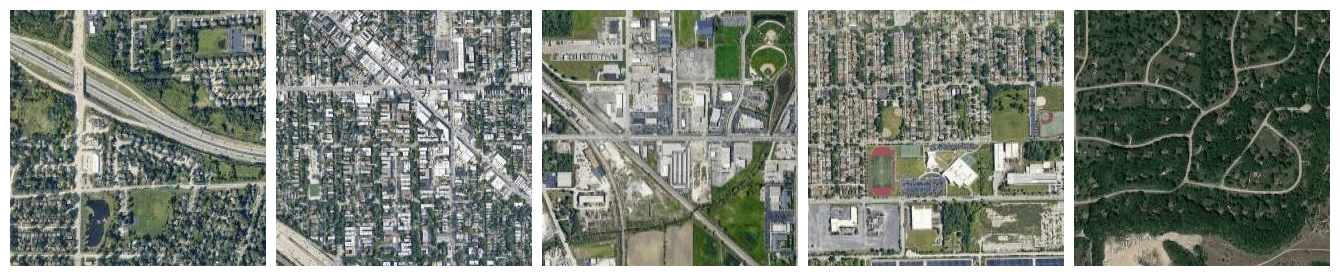}
    \caption{Image samples of cluster centers}
    \label{fig:cluster_imsample} 
\end{figure}

Using K-Means, we identify five clusters with distinct spatial, sociodemographic, and land use patterns. Figure \ref{fig:cluster_imsample} shows the image samples from each cluster. The first cluster represents a suburban town with substantive residential areas, which are cut through by a major highway. The second cluster represents the high-density neighborhoods in the downtown area. The third cluster represents a suburban town center with a more industrial presence. 
The fourth mixes the high-density urban region with some public facilities of large footprints. The fifth cluster is a typical suburban region with very low density and large green space. 

\begin{figure}[ht!]
    \centering
    \subfloat[Entire study area]{
    \includegraphics[width=0.4\linewidth]{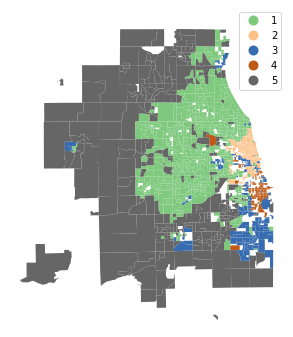}}
    \subfloat[Cook county]{
    \vspace{20pt}
    \includegraphics[width=0.3\linewidth]{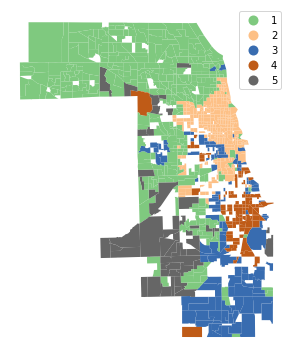}}
    \caption{Spatial distribution of clusters}
    \label{fig:latent_cluster}
\end{figure}

Although spatial information is not explicitly used as inputs, the five clusters can reveal spatial clusters, such as Chicago's urban vs. suburban regions, as shown in Figure \ref{fig:latent_cluster}. The first cluster in northwest Chicago represents the suburban residential regions. The second cluster identifies downtown Chicago. The third and fourth clusters represent southern Chicago, with the former being closer to downtown. The fifth cluster represents the outskirts surrounding Cook County. 

\begin{figure}[ht!]
    \centering
    \subfloat[Density Distribution]{
    \vspace{-12pt}
    \includegraphics[width=0.3\linewidth]{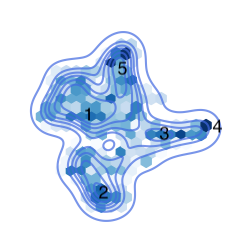}}
    \subfloat[Income per capita]{
    \includegraphics[width=0.3\linewidth]{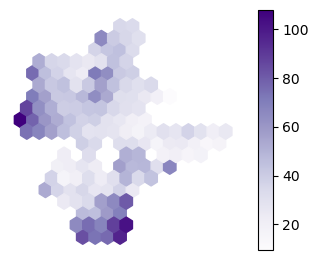}}
    \subfloat[Population (1k ppl/km$^2$)]{
    \includegraphics[width=0.3\linewidth]{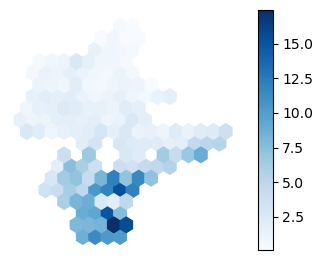}}\\
    \subfloat[Pct Adults]{
    \includegraphics[width=0.3\linewidth]{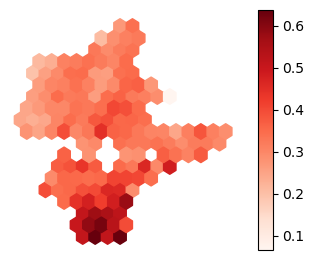}}
    \subfloat[Pct Col Grad]{
    \includegraphics[width=0.3\linewidth]{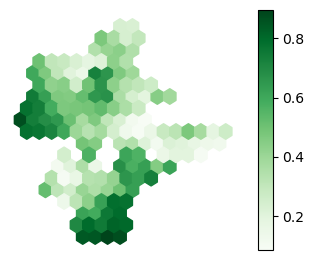}}
    \subfloat[Avg TT to Work (min)]{
    \includegraphics[width=0.3\linewidth]{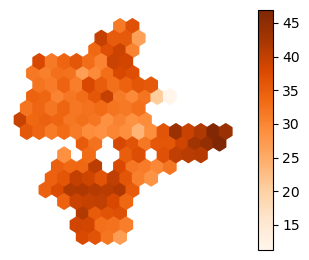}}
    \caption{tSNE visualization of latent space}
    \label{fig:tsne_viz}
\end{figure}

Using tSNE, the latent space could be simplified as a 2D visualization, which can demonstrate its association with the continuous sociodemographic variables. The five clusters are marked in the latent space in Figure \ref{fig:tsne_viz}a, and the continuous sociodemographics in Figure \ref{fig:tsne_viz}b-f. Such continuous patterns in sociodemographics provide additional information to the discrete clusters. For example, the sociodemographics of the first cluster (north) correspond to the high-income and high-education people. The second cluster (downtown) has high income, high population density, a high percentage of adults and college graduates, and moderate to low commuting time. The sociodemographics of the third and fourth (south) clusters correspond to the low income, low population density, low proportions of college graduates, and long commuting time. The last cluster (outskirts) has the lowest density without much sociodemographic richness. 

\subsection{Deriving economic information for generated satellite imagery}
\label{sec:results_economic_interpretation}
\begin{figure}[ht!]
    \centering
    \includegraphics[height=0.2\linewidth]{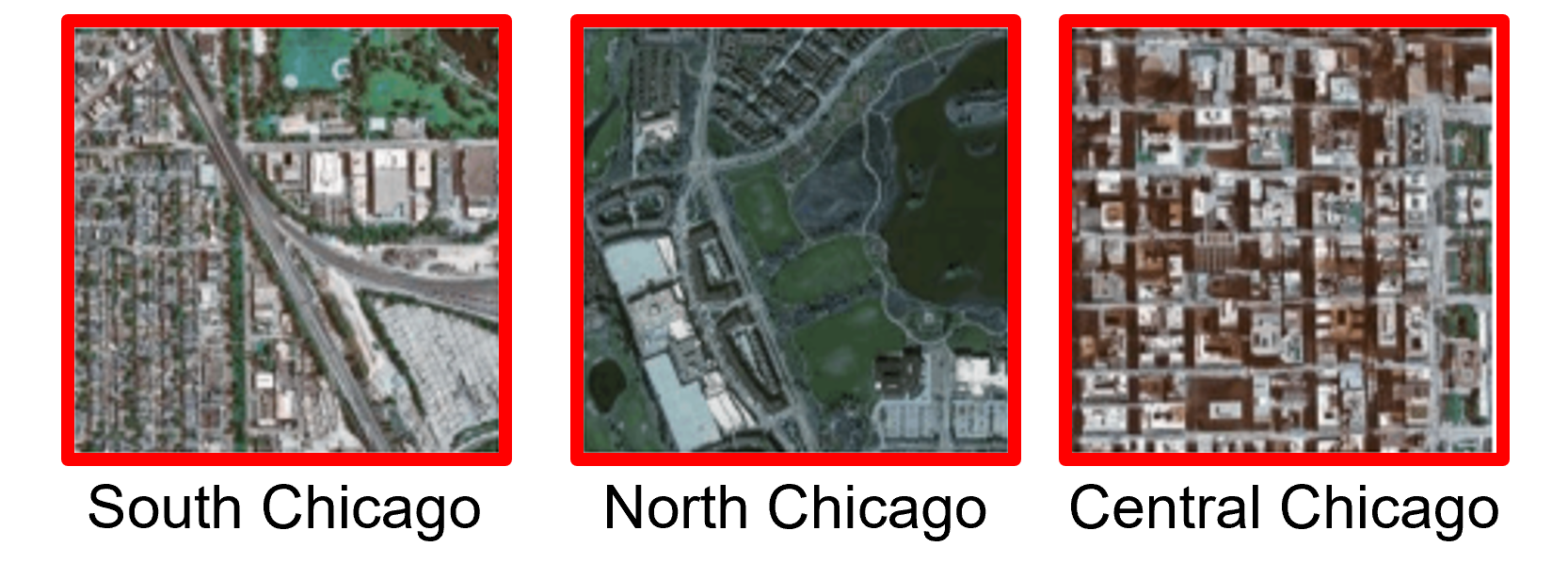}
    \caption{Satellite images of three census tracts} 
    \label{fig:image_complex_context}
\end{figure}

The latent space in DHMs enables us to generate new satellite images and derive economic information from them. The process will be demonstrated by starting with three existing images in Figure \ref{fig:image_complex_context}, which represent the census tracts in the South, North, and Central Chicago areas, the sociodemographics of which are shown in Table \ref{tab:representative_tracts}. The first image presents a mixed urban texture with a major highway in southern Chicago. This area is associated with low income, low education, and long commute times. The other two are located in the north suburban and central Chicago. The north region represents the high-income and low-density areas, while the central region is denser and relatively wealthy. Northern Chicago is represented by a typical suburban pattern with curved roads and low-density buildings, while central Chicago by small-scale and grid-shaped building blocks. The latent variables of the three census tracts are denoted as $z_s$, $z_n$, and $z_c$, and their corresponding images as $I_s$, $I_n$, and $I_c$.

\begin{table}[ht!]
    \centering
    \caption{Sociodemographics of three census tracts}
    \resizebox{\linewidth}{!}{
    \begin{tabular}{l|>{\centering}p{0.15\textwidth}|>{\centering}p{0.12\textwidth}|>{\centering}p{0.15\textwidth}|>{\centering}p{0.14\textwidth}|>{\centering}p{0.10\textwidth}|>{\centering}p{0.10\textwidth}|c}
        \toprule
        Census Tract & Pop Density (k-ppl/$km^2$) & \% College Grad & Avg Time to Work (min)& Income (10k/capita) & Auto & PT & Active \\
        \toprule
        \textbf{S}outh & 2.5 & 10.7\% & 33.6 & 12.1 & 69.9\% & 15.4\% & 5.17\% \\
        \midrule
        \textbf{N}orth & 1.0 & 78.2\% & 37.1 & 90.8 & 86.6\% & 5.1\% & 6.5\% \\
        \midrule
        \textbf{C}entral & 6.2 & 75.9\% & 28.0 & 70.8 & 12.0\% & 46.7\% & 38.4\% \\
        \bottomrule
    \end{tabular}}
    \label{tab:representative_tracts}
\end{table}

\subsubsection{Deriving economic information with one-directional image generation}
\begin{figure}[ht!]
    \centering
    \resizebox{1.0\linewidth}{!}{
    \includegraphics{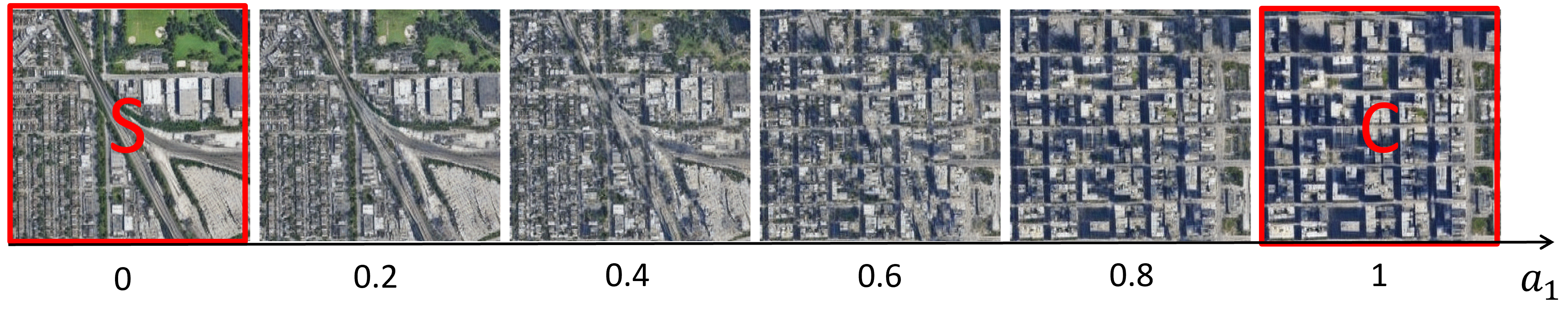}}
    \caption{One-directional image generation}
    \label{fig:image_interpolation}
\end{figure}

New satellite images can be generated using latent variable $\tilde{z}$ through the decoder $\tilde{I} = D_\psi({\tilde{z}})$. Specifically, the latent variable $\tilde{z}$ is created by adding a directional vector ${u}$ to the source latent variable ${z_s}$: $\tilde{z} = z_s + a_1 u$, in which $a_1$ ranging from zero to one in 0.2 increments, and $u$ is the direction between the target and source image $u=z_c-z_s$ in the latent space. As a result, a new satellite image $\tilde{I}$ can be generated with every $\tilde{z}$ as shown in Figure \ref{fig:image_interpolation}. 

\begin{figure}[ht!]
    \centering
    \includegraphics[height=0.15\linewidth]{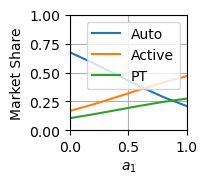}
    \includegraphics[height=0.15\linewidth]{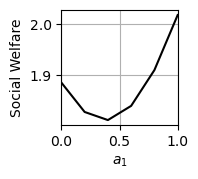}
    \includegraphics[height=0.15\linewidth]{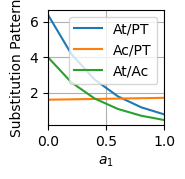}
    \includegraphics[height=0.15\linewidth]{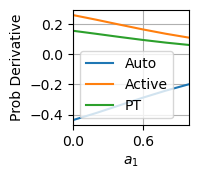}
    \includegraphics[height=0.15\linewidth]{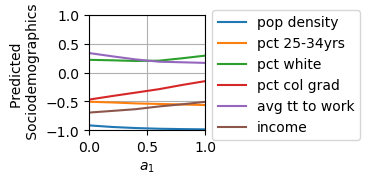}
    \caption{Economic information of images generated along one direction (South to Central Chicago: $\tilde{z} = z_s + a_1(z_c-z_s)$)}
    \label{fig:econ_interpolation}
\end{figure}

For a generated satellite image, its corresponding socioeconomic information can be derived through the DHM framework, including travel behavior $g(\tilde{z})$, utility values $V_{nk}(\tilde{z})$, and sociodemographics $F_\omega(\tilde{z})$. Figure \ref{fig:econ_interpolation} visualizes how market shares, social welfare, substitution patterns, probability derivatives, and sociodemographics vary with the movement in the latent space indicated by the scale factor $a_1$. The market share of auto mode, computed as $e^{\beta_k'\tilde{z}} / {\sum_{j}e^{\beta_{j}'\tilde{z}}}$, decreases when the latent variable moves from the central to the south Chicago. The social welfare, computed with the logsum form $\frac{1}{\alpha_n}  \log (\sum_{j} e^{\beta_j'\tilde{z}})$, significantly decreases when the utilities of travel modes are roughly equivalent. 
The derivatives of choice probabilities  $\nabla_u P_{nk}$ are marginally decreasing when a specific travel mode dominates the mobility market. Overall, larger $a_1$ values are associated with more white people and college graduates, higher income levels, and lower travel time to work. 


\subsubsection{Deriving economic information with two-directional image generation}
\begin{figure}[ht!]
    \centering
    \resizebox{1.0\linewidth}{!}{
    \includegraphics{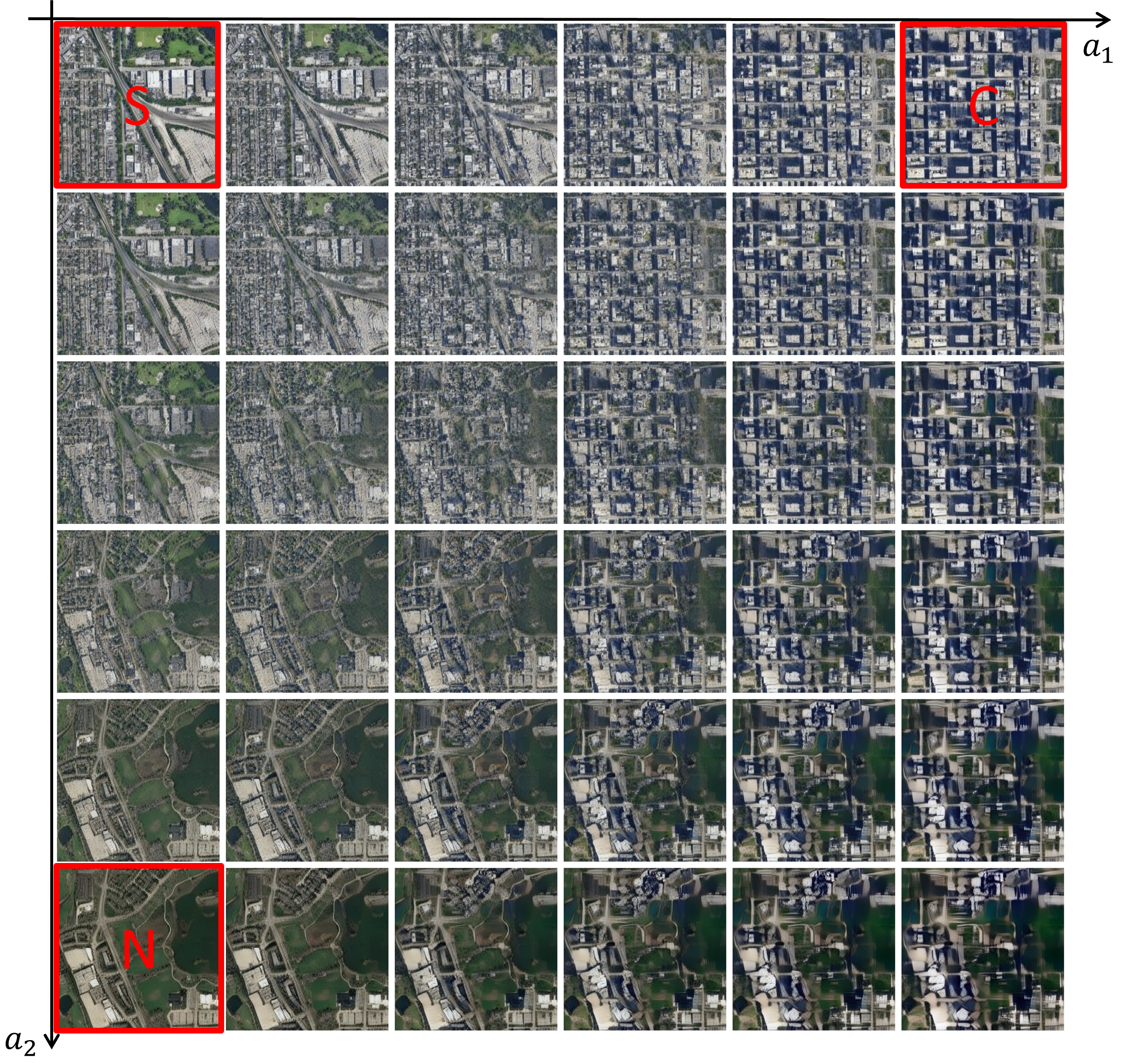}}
    \caption{Two-directional image generation}
    \label{fig:image_complex}
\end{figure}

The one-directional image generation can be extended to a two-directional one using ${\tilde{z}} = a_1 {u_1} + a_2 {u_2} = a_1 ({z_c-z_s}) + a_2 ({z_n-z_s})$. The two directions are defined by $z_c-z_s$ and $z_n-z_s$, representing the movement from the south to north and central Chicago in the latent space. Figure \ref{fig:image_complex} visualizes a 6 $\times$ 6 matrix of the generated images, in which only three images (S, N, and C) exist in reality (Figure \ref{fig:image_complex_context}) while the other 33 images are generated. This two-directional image generation approach generally applies to other satellite images (Appendix III). 

\begin{figure}[ht!]
    \centering
    \subfloat[$a_1=0.6$, $a_2=0$]{\includegraphics[width=0.4\linewidth]{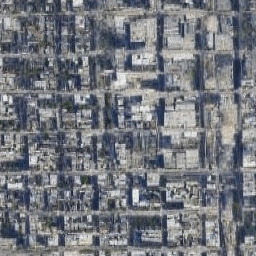}}\qquad
    \subfloat[$a_1=0$, $a_2=0.6$]{\includegraphics[width=0.4\linewidth]{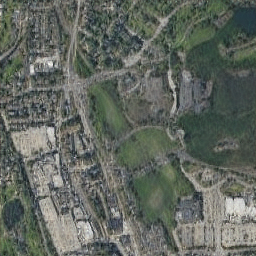}}
    \caption{Two generated satellite images}
    \label{fig:image_interpolation_highlight}
\end{figure}

The generated urban images have high visual quality, as shown in Figure \ref{fig:image_complex} and \ref{fig:image_interpolation_highlight}. Moving from the south to central Chicago, the grid structure and dense buildings gradually dominate the new urban landscape. When $a_1 = 0.6$ and $a_2 = 0.0$ (Figure \ref{fig:image_interpolation_highlight}a), the generated image has a uniform urban grid, resembling central Chicago, with mixed green areas and corridors permeating into this urban area. Moving from the south to north Chicago, the major north-south green belt is gradually replaced by curved suburban roads and low-density buildings. When $a_1 = 0.0$ and $a_2 = 0.6$ (Figure \ref{fig:image_interpolation_highlight}b), the urban landscape retains the curved suburban road networks similar to north Chicago, but with mixed building footprints. 


\begin{figure}[ht!]
    \centering
    \subfloat[Market share - Automobiles]{\includegraphics[width=0.22\linewidth]{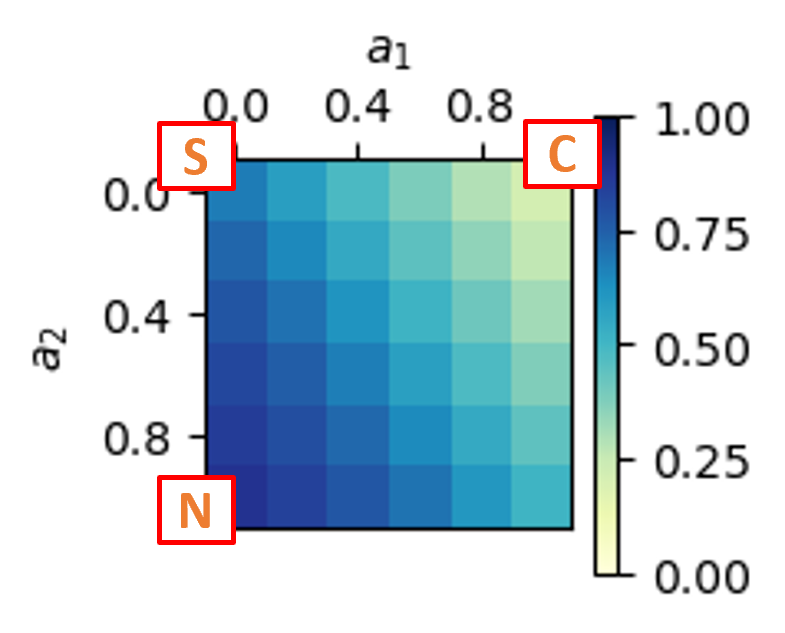} }
    \subfloat[Market share - Active mode]{\includegraphics[width=0.22\linewidth]{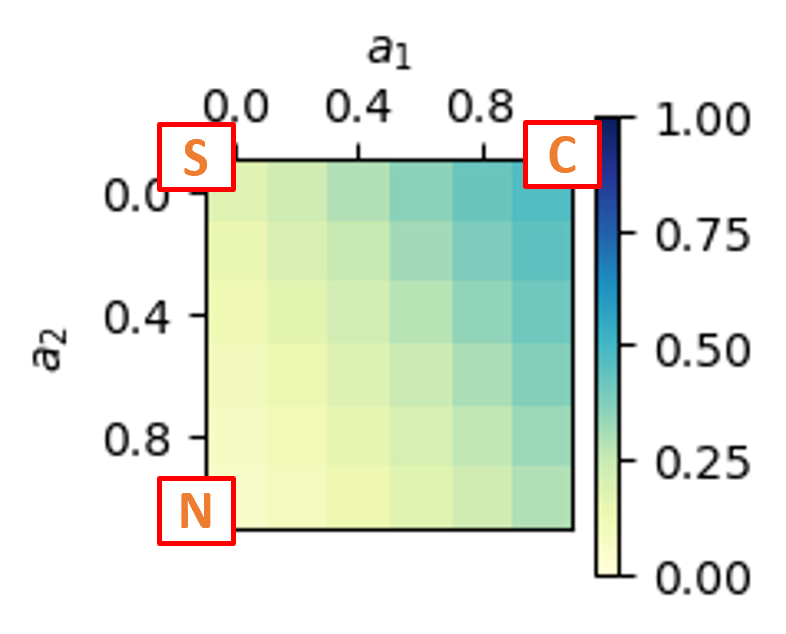}}
    \subfloat[Market share - PT]{\includegraphics[width=0.22\linewidth]{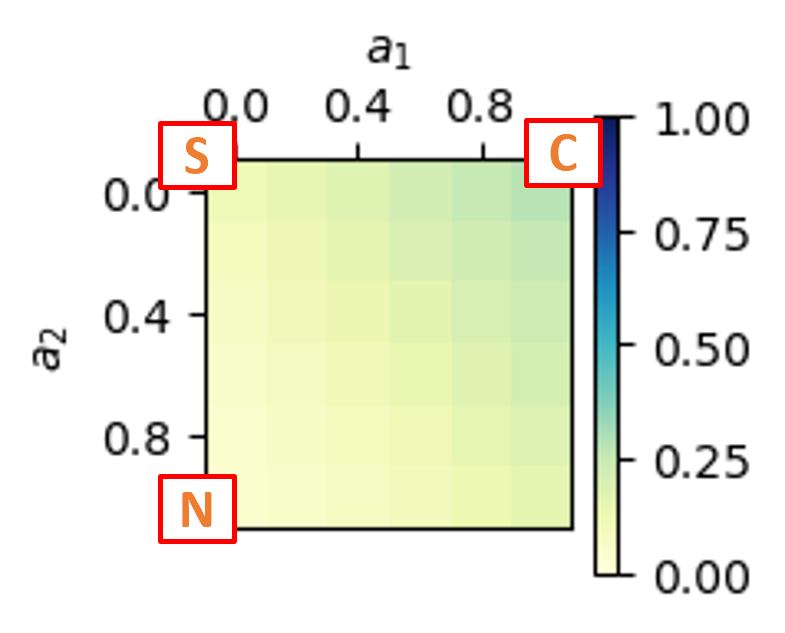} }
    \subfloat[Social welfare]{\includegraphics[width=0.22\linewidth]{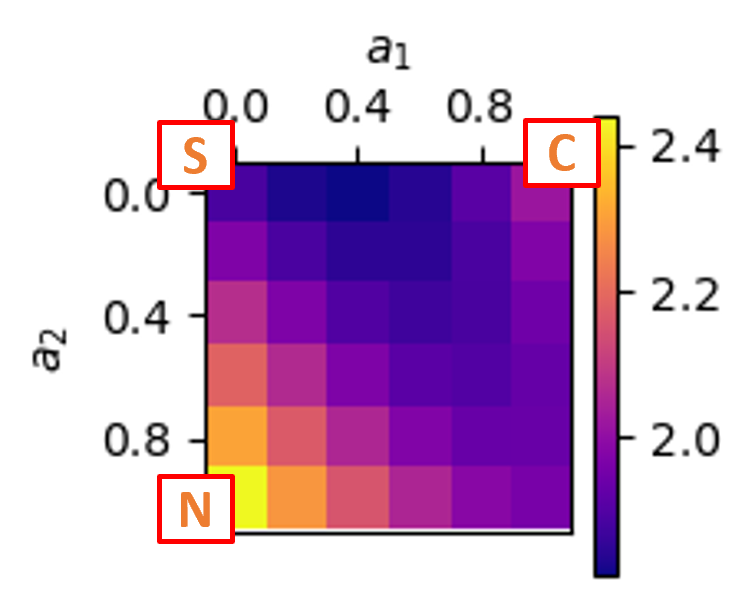}}
    \caption{Economic information of images generated along two directions}
    \label{fig:econ_image_complex}
\end{figure}

Similar to the one-directional examples, the socioeconomic information can be derived for every generated satellite image, as shown in Figure \ref{fig:econ_image_complex}.  The satellite image on row $i$ and column $j$ in Figure \ref{fig:image_complex} corresponds to the market share and social welfare values in the pixel on row $i$ and column $j$ in Figure \ref{fig:econ_image_complex}. Such matrix plots in Figure \ref{fig:econ_image_complex} can provide a holistic view of the two-directional interactions. For example, public transit usage does not change significantly by moving from southern to northern Chicago, but varies by moving from southern to central Chicago. Beyond market shares and social welfare, it is also feasible to compute other economic parameters for the generated image $\tilde{I}$. Here we skip further details and leave them for future studies. 

\section{Conclusion}
\label{sec:conclusion}
Travel decisions can be influenced by the factors represented by either numeric data or urban imagery. Although classical demand modeling cannot effectively process urban imagery, it becomes imperative to expand on this capacity due to the explosion of unstructured data. This study proposes the DHM framework to integrate numeric and satellite imagery by combining the analytical capacity of the demand modeling and the computational capacity of the DNNs. After empirically examining its performance, we demonstrate three major findings of DHMs, which correspond to three contributions of this research. 

First, this study demonstrates the complementarity of data and model. The data complementarity is demonstrated by the higher performance of the DHMs combining numeric and imagery data over the benchmark models. The model complementarity is demonstrated by integrating classical demand modeling and deep learning through the latent space in the DHMs. Technically, the complementarity is achieved by the SAE design. Unlike the simple concatenation, SAEs combine the imagery and sociodemographics through the supervised sociodemographics for AE. Second, this study enriches the hybrid demand models with deep hybrid models by constructing a latent space from the satellite imagery using deep architectures. This latent space contains meaningful social and spatial characteristics, although its interpretation is no longer as straightforward as the classical hybrid demand models. Third, DHMs offer one method to compute economic information for the generated satellite imagery. The high-quality images are generated using the linearly transformed latent variables through the AE in stable diffusion models. The economic information related to the generated urban images (e.g. mode shares and social welfare) can be computed because the DHMs successfully associate sociodemographics, travel behavioral outputs, and satellite imagery through the latent space.

Our framework is hybrid for five reasons. Three reasons are already fully elaborated: it is hybrid because it integrates numeric and imagery data, combines classical demand modeling and deep learning, and enriches the classical hybrid model family. Two other reasons, relatively less elaborated so far, are on the machine learning side. The DHM is hybrid because it mixes supervised and unsupervised learning: the vertical axis of DHMs in Figure \ref{fig:dhm_framework} represents supervised learning, while the horizontal autoencoder represents unsupervised learning. It also mixes the discriminative and generative methods. The decoder generates new satellite imagery while the behavioral prediction is discriminative. The two perspectives are explained in other machine learning studies that adopted the hybrid generative-discriminative or supervised-unsupervised methods \cite{Lasserre2006PrincipledModels, Grabner2007, Donahue2016, Lin2019}. 


The DHM framework still has many limitations. The DHMs leverage the unique capacity of the deep autoencoders and stable diffusion models to process urban imagery. However, this strength is inevitably associated with the challenges in a highly complex neural network. Although deep learning can minimize errors effectively, the final solution by no means is a globally optimal solution. Neural networks converge to various local minima, and this non-identification issue is still a pending challenge in deep learning research. The economic information computed from the generated satellite imagery is a numerical approximation, as opposed to the analytical solutions from the classical parameter-based discrete choice models. The latent space in DHMs is no longer constrained to a small number of latent variables, limiting its interpretability in the classical parameter-based sense. It is also ambiguous how to guarantee stability and robustness in such a high-dimensional latent space. The DHMs achieve relatively high predictive performance, but their transferability could be limited. It is unclear whether the models can still achieve high performance when they are trained in one context and used in another. This study uses the autoencoder in the state-of-the-art stable diffusion for image generation, but the quality of image generation can always be further improved. Meanwhile, the U-Net architecture in Models 5-6 is computationally expensive, thus limiting its replicability for future researchers. Regarding future work, the DHMs combine two purposes: representation learning for prediction and image regeneration for image-based story-telling. But it might not be easy to improve two purposes simultaneously, so future studies can focus on achieving one purpose without sacrificing the other. Empirically, researchers could explore how to use satellite imagery in other contexts. For example, sociodemographics are often considered exogenous variables for energy consumption, health, and air pollution, so future research could combine satellite imagery with sociodemographics to analyze these factors using the DHM framework. 

\section*{CRediT authorship contribution statement}
\noindent
\textbf{Qingyi Wang:} Conceptualization, Methodology, Software, Data Curation, Writing - Original Draft, Writing - Review \& Editing, Visualization; 
\textbf{Shenhao Wang:} Conceptualization, Methodology, Writing - Original Draft, Writing - Review \& Editing, Supervision, Project administration, Funding acquisition; 
\textbf{Yunhan Zheng}: Writing - Review \& Editing; 
\textbf{Hongzhou Lin}: Conceptualization; 
\textbf{Xiaohu Zhang}: Data Curation; 
\textbf{Jinhua Zhao}: Supervision, Funding acquisition; 
\textbf{Joan Walker}: Supervision. The authors declare no conflict of interest.

\section*{Acknowledgements}
\noindent
This material is based upon work supported by the U.S. Department of Energy’s Office of Energy Efficiency and Renewable Energy (EERE) under the Vehicle Technology Program Award Number DE-EE0009211. The views expressed herein do not necessarily represent the views of the U.S. Department of Energy or the United States Government. The authors are also grateful for the early RA support from Rachel Luo and Jason Lu, and David Bau for insightful discussions.

\printbibliography
\newpage

\setcounter{table}{0}
\renewcommand{\thetable}{A\arabic{table}}

\setcounter{figure}{0}
\renewcommand{\thefigure}{A\arabic{figure}}

\section*{Appendix I. Specific model design in disaggregate analysis}
\label{appendix:specific_design}
To simplify the notation used in the manuscript, we only distinguish between imagery latent variables and numeric variables in our discussion. In fact, there are three types of inputs to the behavioral predictors: imagery latent variables, sociodemographics, and alternative-specific variables. Both sociodemographics and alternative-specific variables are numeric but they enter the utility function slightly differently. In this paper, we used $z_n$ to represent the imagery latent space, $x^{sd}$ and $x^{alt}$ represent the sociodemographic and alternative-specific attributes, respectively. Combining all three types of attributes, we formulate the utility function as follows:
\begin{equation}
    V_{nk} = \beta^{im'}_{k} z_n + \beta^{sd'}_{k} x^{sd}_n  + \beta^{alt'} x^{alt}_{nk} 
\end{equation}

\noindent where  $\beta^{im'}_{k}$,  $\beta^{sd'}_{k}$, and $\beta^{alt'}$ are vector coefficients of urban imagery latent space, sociodemographics, and alternative-specific attributes, respectively. We assume that the alternative-specific variables share the same coefficients in the utility of all alternatives. Both imagery and sociodemographics are not alternative-specific, therefore the coefficients $\beta^{im'}_{k}$,  $\beta^{sd'}_{k}$ are alternative-specific to distinguish the effect these variables have on different alternatives.


In Section \ref{sec:methods_econ_info}, we also discussed the directional probability derivatives for the imagery latent variable $z_n$. We could also calculate probability derivatives for the other two types of variables. Both sociodemographic and alternative-specific variables are standard numeric variables and many references exist on the how to get the probability derivatives. We only summarize the results below.

For alternative-specific attributes, the probability derivative, which is the change in probability that a trip $n$ uses alternative $k$ with respect to changes in variable $i$ is simply
\begin{equation}
    \frac{\partial P_{nk}}{\partial x^{alt}_{nki}} = \frac{\partial V_{nk}}{\partial x^{alt}_{nki}}P_{nk}(1-P_{nk}) = \beta^{alt}_{i}P_{nk}(1-P_{nk})
\end{equation}
Sociodemographic variables enter the utility of all alternatives with different $\beta$ parameters. Therefore the change in probability that a trip $n$ uses alternative $k$ with respect to changes in variable $i$ is
\begin{equation}
    \frac{\partial P_{nk}}{\partial x^{sd}_{ni}} =  \sum_{k'} \frac{\partial P_{nk}}{\partial V_{nk'}} \frac{\partial V_{nk'}}{\partial x^{sd}_{ni} } = \beta^{sd}_{ki}P_{nk}(1-P_{nk}) - \sum_{k'}{\beta^{sd}_{k'i}P_{nk}P_{nk'}}
\end{equation}
We can compute a directional gradient of choice probabilities regarding the imagery latent space, which resembles but also differs from the marginal effects $\beta$ in the classical demand modeling. The formula of the directional gradient is  
\begin{equation}
    \nabla_u P_{nk}({z}) = {u} \cdot \nabla P_{nk}({z})  = 
    \sum_{k'} \frac{\partial P_{nk}}{\partial V_{nk'}} \nabla_u V_{nk'}({z_n}) = {u} \cdot ({\beta^{im}_{k}}P_{nk}(1-P_{nk}) - \sum_{k'}{{\beta^{im}_{k'}}P_{nk}P_{nk'}})
\end{equation}

\newpage
\section*{Appendix II. Effects of mixing and sparsity hyperparameters on prediction and image reconstruction}
The mixing hyperparameter $\lambda$ affects the performances in behavioral predictors by controlling the degree of information mix in the latent space. The degree of information mix can be understood by the trade-off between retaining the richness in image reconstruction and encoding sociodemographic information. 
Table \ref{tab:reg_lambda} shows the predictive performance under different $\lambda$ values. Since $\lambda=0$ represents an autoencoder without sociodemographic information, the non-zero $\lambda$ values in Table \ref{tab:reg_lambda} imply different degrees of information mixing. 
The improvement of performance can be quite significant by optimizing $\lambda$. 
In general, the mixing hyperparameter should not be set too small or too large. Among the five values tested, $\lambda=0.7$ and $0.9$ achieve the best performance in Panel 1, while $\lambda = 0.5$ and $0.7$ achieve the best performance in Panels 2 and 3. Overall $\lambda=0.7$ achieves near-optimal performance across all panels and therefore it is used in the subsequent analyses. This observation corroborates the claim that the two data sources are complementary and achieves the highest predictive performance when they are properly mixed.

\begin{table}[ht!]
    \centering
    \caption{Effects of the mixing hyperparameter $\lambda$ in Model 4}
    \resizebox{\linewidth}{!}{
    \begin{tabular}{l|c|c|c|c|c}
        \toprule
        \multicolumn{1}{c|}{\textbf{The mixing hyperparameter $\lambda$}} & $0.1$  & $0.3$ & $0.5$ & $0.7$ & $0.9$ \\
        \midrule
        \multicolumn{5}{l}{\textit{Panel 1: Aggregate Mode Choice - Linear Regression}} \\
        \midrule
        Auto ($R^2$) & 0.621/0.593 & 0.635/0.592 & 0.688/0.633 & \textbf{0.676/0.640} & 0.688/0.637\\
        \midrule
        Active ($R^2$) & 0.501/0.474 & 0.549/0.480 & 0.582/0.528 & 0.565/0.524 & \textbf{0.589/0.533}\\
        \midrule
        PT ($R^2$) & 0.463/0.444 & 0.514/0.473 & 0.538/0.473 & 0.520/0.483 & \textbf{0.519/0.488}\\
        \midrule

        \multicolumn{5}{l}{\textit{Panel 2: Aggregate Mode Choice - Multinomial Regression}} \\
        \midrule
        KL Loss & 0.139/0.147 & 0.135/0.148 & 0.110/0.141 & \textbf{0.101/0.140} & 0.104/0.145 \\
        Auto ($R^2$) & 0.606/0.581 & 0.607/0.559 & \textbf{0.706/0.605} & 0.737/0.604 & 0.734/0.590\\
        Active ($R^2$) & 0.484/0.433 & 0.493/0.435 & 0.630/0.472 & \textbf{0.657/0.485} & 0.668/0.469\\
        PT ($R^2$) & 0.437/0.393 & 0.481/0.406 & \textbf{0.608/0.421} & 0.674/0.421 & 0.650/0.363\\
        \midrule 
        
        \multicolumn{5}{l}{\textit{Panel 3: Disaggregate Mode Choice}} \\
        \midrule
        CE Loss & 0.376/0.402 & 0.378/0.412 & 0.371/0.406 & \textbf{0.373/0.404} & 0.461/0.467\\
        Accuracy & 0.870/0.858 & 0.870/0.856 & \textbf{0.872/0.861} & 0.871/0.856 & 0.843/0.831 \\
        \bottomrule
    \end{tabular}
    }
    \label{tab:reg_lambda}
\end{table}

\begin{figure}[ht!]
    \centering
    \includegraphics[width=\linewidth]{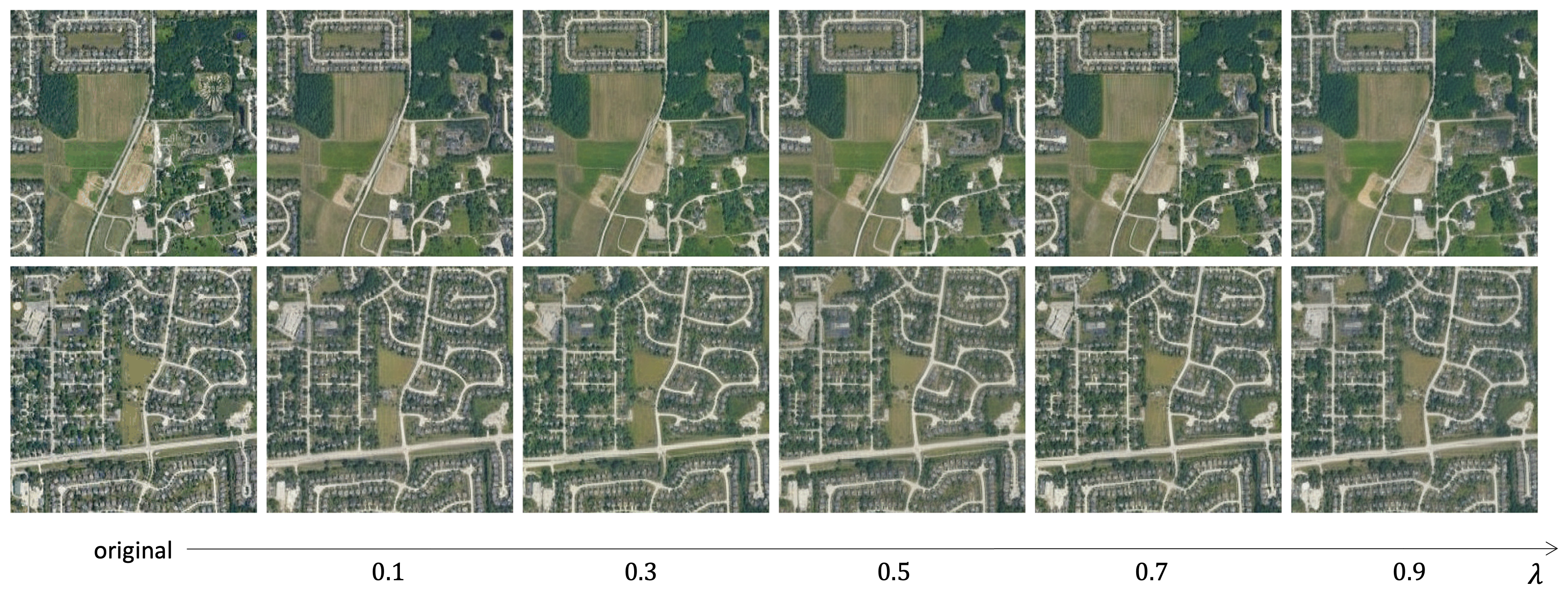}
    \caption{Quality of image reconstruction with various $\lambda$ values}
    \label{fig:img_recon}
\end{figure}

The reconstruction quality between different $\lambda$'s does not differ much because of the various image quality enhancement terms (GAN, KL, LPIPS) in the formulation. Figure \ref{fig:img_recon} shows two sample image reconstructions with respect to changing $\lambda$. While smaller $\lambda$'s provide more textured details, all $\lambda$'s achieve high-quality image reconstructions.

Besides the mixing hyperparameter $\lambda$, the choice of the sparsity hyperparameter $\theta$ is also key to high model performance. The importance of $\theta$ can be observed by Table \ref{tab:reg_theta}, in which $\lambda$ is fixed to 0.7, and we observe that a non-zero $\theta$ value is always necessary in all our models and tasks to achieve the highest testing performance. Unlike $\lambda$, the range of $\theta$ values is explored on an exponential scale, and the search range heavily depends on the type of behavioral predictor. Regardless of the details, a high-dimensional latent space is often necessary for encoding urban imagery, and to enhance the generalizability of the latent space predictive power, the sparsity control through the hyperparameter $\theta$ is critical. 

\begin{table}[ht!]
    \centering
    \caption{Effects of the sparsity hyperparameter $\theta$ in Model 4}
    \resizebox{0.85\linewidth}{!}{
    \begin{tabular}{l|c|c|c|c|c}
        \toprule
        \multicolumn{5}{l}{\textit{Panel 1: Aggregate Mode Choice - Linear Regression}} \\
        \midrule
        $\theta$ & $1e^{-4}$ & $1e^{-3}$ & $1e^{-2}$ & $1e^{-1}$ & $1e^0$\\
        \midrule
        Auto ($R^2$) & 0.996/0.183 & 0.894/0.532 & \textbf{0.674/0.640} & 0.606/0.594 & 0.222/0.217 \\
        \midrule
        Active ($R^2$) & 0.826/0.425 & 0.592/0.521 & \textbf{0.566/0.524} & 0.553/0.520 & 0.543/0.515  \\
        \midrule
        PT ($R^2$) & 0.716/0.445 & 0.538/0.483 & \textbf{0.520/0.483} & 0.510/0.482 & 0.503/0.480  \\
        \midrule
        \multicolumn{5}{l}{\textit{Panel 2: Aggregate Mode Choice - Multinomial Regression ($\lambda^*=1e^{+3}$)}} \\
        \midrule
        $\theta$ & $1e^{+1}$ & $1e^{+2}$ & $1e^{+3}$ & $1e^{+4}$ & $1e^{+5}$ \\
        \midrule
        KL Loss & 0.099/0.141 & 0.103/0.140 & 0.108/0.134 & \textbf{0.120/0.129} & 0.129/0.133\\ 
        Auto ($R^2$) & 0.748/0.591 & 0.729/0.597 & 0.713/0.620 & \textbf{0.669/0.637} & 0.639/0.620\\
        Active ($R^2$) & 0.663/0.463 & 0.661/0.471 & 0.633/0.501 & \textbf{0.566/0.515} & 0.530/0.505\\
        PT ($R^2$) & 0.705/0.411 & 0.631/0.394 & 0.629/0.451 & \textbf{0.543/0.496} & 0.502/0.489\\
        \midrule 
        
        \multicolumn{5}{l}{\textit{Panel 3: Disaggregate Mode Choice ($\lambda^*=1e^0$)}} \\
        \midrule
        $\theta$ & $1e^{-5}$ & $1e^{-4}$ & $1e^{-3}$ & $1e^{-2}$ & $1e^{-1} $\\
        \midrule
        CE Loss & 0.389/0.415 & 0.397/0.420 & \textbf{0.373/0.404} & 0.376/0.413 & 0.395/0.418 \\
        Accuracy & 0.864/0.854 & 0.861/0.853 & \textbf{0.871/0.856} & 0.870/0.855 & 0.861/0.852 \\
        \bottomrule
    \end{tabular}
    }
    \label{tab:reg_theta}
\end{table}

\newpage
\section*{Appendix III. Two-directional imagery regeneration}
\begin{figure}[ht!]
    \centering
    \resizebox{1.0\linewidth}{!}{    \includegraphics{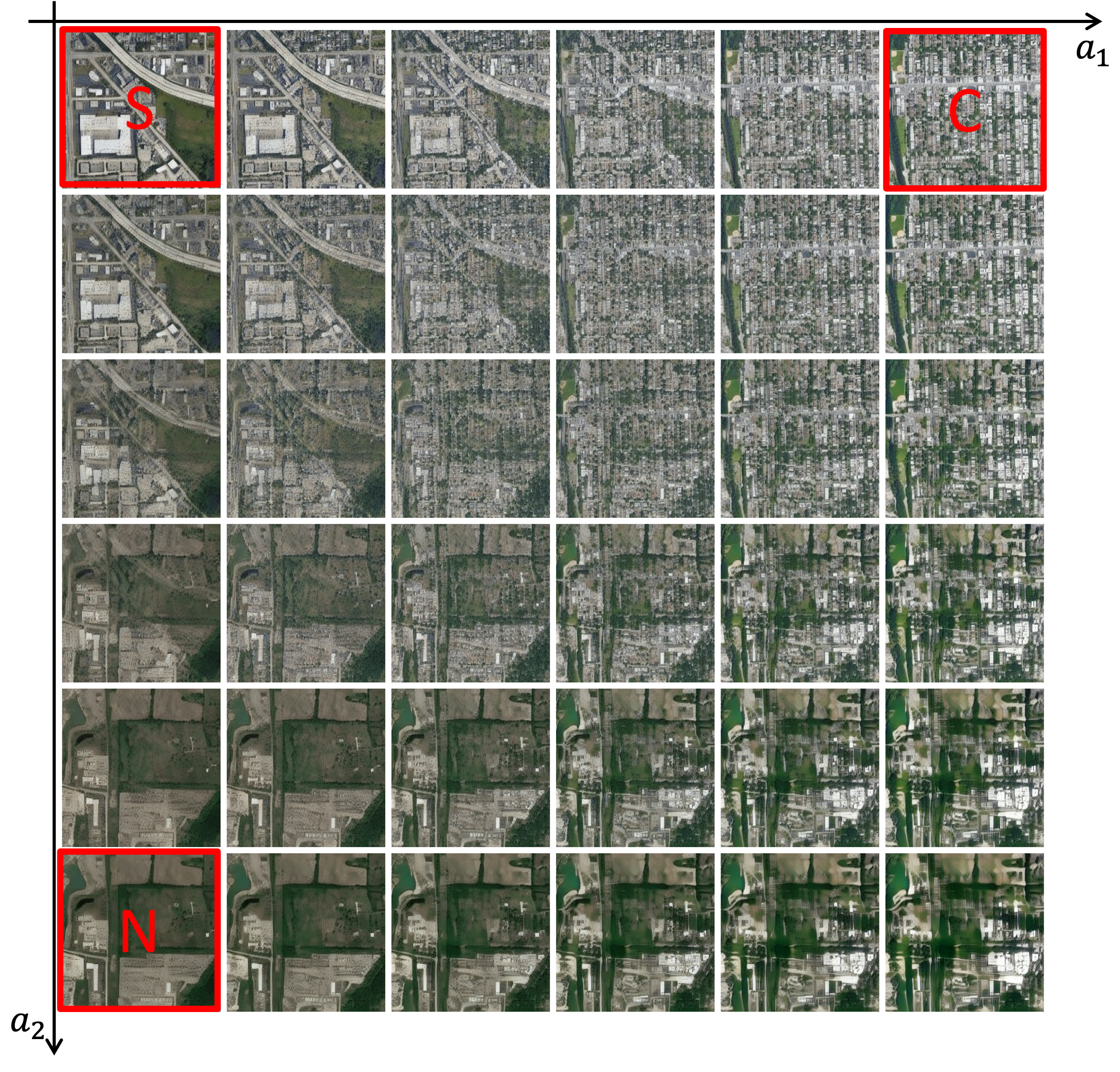}}
    \caption{Another example of two-directional image regeneration}
\end{figure}

\noindent This figure demonstrates another example of reconstructing satellite images with the 2D directional movements in the latent space. Similar to Figure \ref{fig:image_complex}, we choose three targeting images, as highlighted by the red squares, to generate new urban images by linear interpolation and extrapolation in the latent space. The images appear realistic, and through our framework, they are also associated with socioeconomic and behavioral information. 
\end{document}